\documentclass[10pt,twocolumn,letterpaper]{article}
\usepackage{iccv}
\usepackage{times}
\usepackage{epsfig}
\usepackage{graphicx}
\usepackage{amsmath}
\usepackage{amssymb}
\usepackage{cuted}

\usepackage[pagebackref=true,breaklinks=true,colorlinks,bookmarks=false]{hyperref}

\usepackage[capitalize]{cleveref}
\crefname{section}{Sec.}{Secs.}
\Crefname{section}{Section}{Sections}
\Crefname{table}{Table}{Tables}
\crefname{table}{Tab.}{Tabs.}
\newcommand{\Dclip}{D_{\text{CLIP}}}

\iccvfinalcopy 

\def\iccvPaperID{11325} 

\ificcvfinal\fi

\begin{document}

\title{MetaHead: An Engine to Create Realistic Digital Head}

\author{\large Dingyun Zhang\textsuperscript{1} \quad Chenglai Zhong\textsuperscript{1} \quad Yudong Guo\textsuperscript{2}  \quad Yang Hong\textsuperscript{1} \quad Juyong Zhang\textsuperscript{1} \vspace{1.5 mm}\\
	{\textsuperscript{1}University of Science and Technology of China \quad \textsuperscript{2}Image Derivative Inc}}

\maketitle
\ificcvfinal\thispagestyle{empty}\fi

\begin{abstract}
Collecting and labeling training data is one important step for learning-based methods because the process is time-consuming and biased. For face analysis tasks, although some generative models~\cite{karras2019style,deng2020disentangled,shoshan2021gan,chan2021pi,chan2022efficient,deng2022gram,hong2022headnerf} can be used to generate face data, they can only achieve a subset of generation diversity, reconstruction accuracy, 3D consistency, high-fidelity visual quality, and easy editability. One recent related work is the graphics-based generative method~\cite{wood2021fake}, but it can only render low realism head with high computation cost. In this paper, we propose MetaHead, a unified and full-featured controllable digital head engine, which consists of a controllable head radiance field(MetaHead-F) to super-realistically generate or reconstruct view-consistent 3D controllable digital heads and a generic top-down image generation framework LabelHead to generate digital heads consistent with the given customizable feature labels. Experiments validate that our controllable digital head engine achieves the state-of-the-art generation visual quality and reconstruction accuracy. Moreover, the generated labeled data can assist real training data and significantly surpass the labeled data generated by graphics-based methods in terms of training effect. The project page is available at: \href{https://ustc3dv.github.io/MetaHead/}{https://ustc3dv.github.io/MetaHead/}.
\end{abstract}

\section{Introduction}
Generation and reconstruction of digital human are in increasing demand in virtual reality~\cite{schuemie2001research}, game and movie character production, and metaverse(shared virtual 3D world)~\cite{lee2021all}, as digital human with controllable geometry and appearance could bring realistic visual experience.
On the other hand, semantic labels of heads such as 2D/3D landmarks~\cite{zhu2012face}, eye gaze angle~\cite{zhang2015appearance} and hair color are valuable and necessary precondition for many facial analysis tasks~\cite{zhao2003face,cowie2001emotion,thies2016face2face,wang2002study}, and have various applications like face alignment~\cite{jin2017face} and registration, facial expression analysis~\cite{loven2012revisiting}, video conferencing~\cite{criminisi2003gaze}, human computer interface~\cite{hartson1989human}. Large well-labeled training data~\cite{hastie2009overview} is the key to the success of learning-based models. The more precise the labels, the better the learning-based models will perform. Labeled instances however are often difficult, expensive, or time-consuming to obtain, as they require the efforts of experienced human annotators. In addition, most of the data collected by the existing head label database is under medium pose and illumination, and there are very little challenging big pose and dark-illumination data.
Therefore, learning-based models still have a lot of room for improvement in challenging scenarios and generalization.

A digital head engine is an all-in-one model that can reconstruct and control digital heads, and generate digital heads consistent with the customizable head feature labels. So far, little work has been done on the unified engine~\cite{wood2021fake}, despite the growing promise and demand for it. Graphics-based methods~\cite{wood2021fake} can generate heads consistent with specified labels. However, there exists obvious gap between the synthesized texture distribution and real texture distribution, which results that the final synthesized heads look like cartoon images.

Recently, learning-based models get more and more attentions. Among them, heads generated by early 2D-based GAN models~\cite{karras2019style,deng2020disentangled,shoshan2021gan} lack view consistency. The 3D generative adversarial structure models based on neural radiance field (NeRF) representation~\cite{deng2022gram,chan2022efficient} can improve the view consistency of head geometry and appearance. However, since the latent space of GAN is hard to semantically manipulate with disentanglement, they do not allow flexible head control of the output images. The model introduced with 3DMM prior~\cite{hong2022headnerf} can reconstruct the controllable heads(it is not a generative model), but its reconstruction quality is poor on photorealism and clarity. When changing the viewing angle, the heads obtained by the above models will have hair and teeth flickering phenomenon, which destroys the visual effect. In addition, their generation or control of heads only covers medium poses and simple expressions (smile and mouth opening) due to their limited training data and attributes controlling disentanglement.

We propose MetaHead, a super-realistic controllable head engine, which realizes the reconstruction, control, and generation of heads consistent with the given labels. A naive baseline solution is to represent the 3D head scene based on NeRF, and use the decoupled 3DMM coefficients as shape and appearance prior conditions to input NeRF, and utilize real 2D images as supervision, similar to ~\cite{hong2022headnerf}. However, this end-to-end solution is actually the decoder structure of the Auto-encoder, and the generated figures are very blurry~\cite{larsen2016autoencoding}. At the same time, the 3DMM coefficients are based on the Base Face Model(BFM)~\cite{paysan20093d}, and the identity bases are sparse(built
from very limited 3D scans). The implicit semantics of identity and expression coefficients are also not good at direct 3D control over face geometry during volume learning procedure.

In MetaHead, we propose a controllable head radiance field(MetaHead-F) by designing a strategy to combine the decoder with the pre-trained GAN generator. This end-to-end structure ensures that the disorganized latent space can be decoupled via learning. GAN does not use point-wise loss during pre-training but distribution-matched GAN loss, so the convergence point is close to the data manifold surface~\cite{DBLP:journals/corr/BergmannJV17} and thus MetaHead-F can generate visually clear human heads. The generator unit inspired by StyleGAN3~\cite{karras2021alias} can suppress aliasing information, in which we designed a hierarchical structure attention module to solve the chronic problem of hair and teeth flickering when the viewing angle changes. Furthermore, we add a parallel semantic network to implicitly learn hair and mouth contours. We use face recognition features and the 3D landmarks in datum space as the prior condition signal of the head geometry to directly input into MetaHead-F, so that MetaHead-F can accurately and effectively reconstruct, generate and control the head identity and expression, covering challenging poses and expressions. 3DMM texture coefficients and spherical harmonic illumination~\cite{green2003spherical} coefficients are input to MetaHead-F as appearance condition signals.

MetaHead-F is the main body of the MetaHead engine model. We also propose LabelHead, a top-down paradigm in which MetaHead could generate heads consistent with the customizable head labels. The specified head features would be embedded in the latent space of MetaHead-F, leading it to a huge design space. We can assign label values to each feature to generate head images with various feature labels. We took landmark feature(see \cref{subsec:Qualitative Evaluation on LabelHead} and \cref{subsec:Quantitative Evaluation on LabelHead}) and eye gaze angle feature(see \cref{subsec:eye gaze angle}) as examples to verify that the labeled data generated by MetaHead can remarkably assist real data and significantly surpass the labeled data synthesized by graphics-based methods. 

It is noteworthy that when reconstructing the heads, MetaHead-F would perform backward fitting on labels of head features. Therefore, MetaHead can bidirectionally inference labels of features such as landmark coordinates, eye gaze angle, hair color and so on that are difficult to annotate before. In summary, our main contributions include:
\begin{itemize}
\item We propose MetaHead, the unprecedented learning-based super-realistic controllable head engine, which combines and realizes the head generation, reconstruction, 3D control, and generating heads consistent with the given head labels. Furthermore, it can also bottom-up estimate the labels of head features bidirectionally.
\item We propose a controllable head radiance field (MetaHead-F) to generate or reconstruct view-consistent 3D controllable digital heads, which unifies the advantages of end-to-end and GAN structures, and achieves state-of-the-art in head reconstruction accuracy, control accuracy and generation visual quality.
\item We propose a generic top-down image generation framework LabelHead to generate heads with the customizable feature labels. It enables synthesizing large amount of labeled head images with various shapes and appearances. The developed framework LabelHead can be applied to any shape-appearance related fields other than human head images.
\end{itemize}

\begin{figure*}[t]
  \centering
   \includegraphics[width=1.0\linewidth]{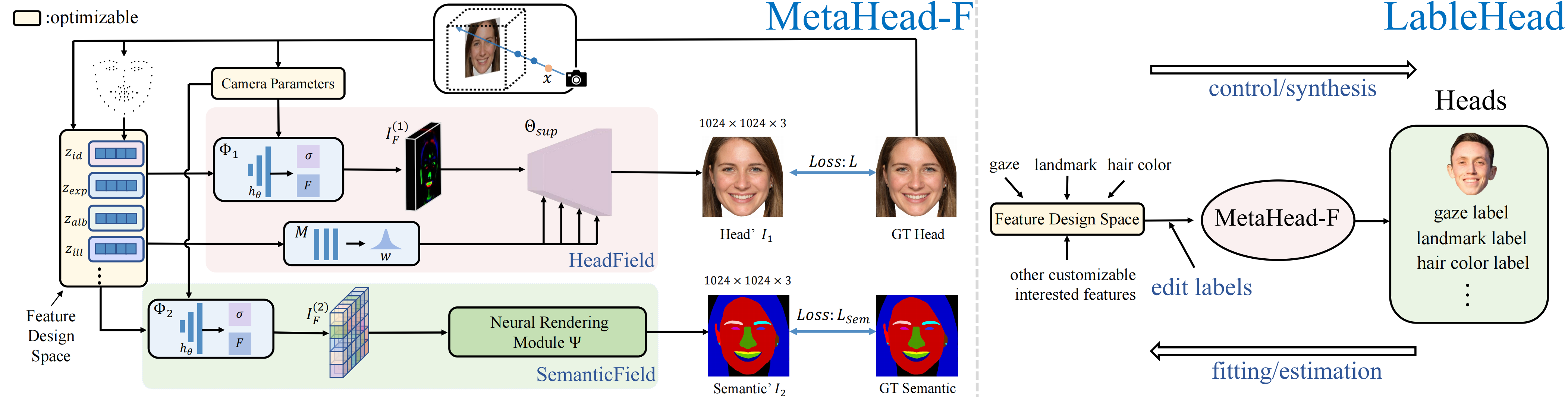}
   \caption{Overview of digital head engine MetaHead. It consists of a controllable head radiance field (MetaHead-F) to super-realistically reconstruct or generate view-consistent 3D controllable digital heads and a generic top-down image generation framework LabelHead to generate heads consistent with the given customizable feature labels.}
   \label{fig:pipeline}
\end{figure*}
\begin{figure}[t]
  \centering
  \includegraphics[width=1.0\linewidth]{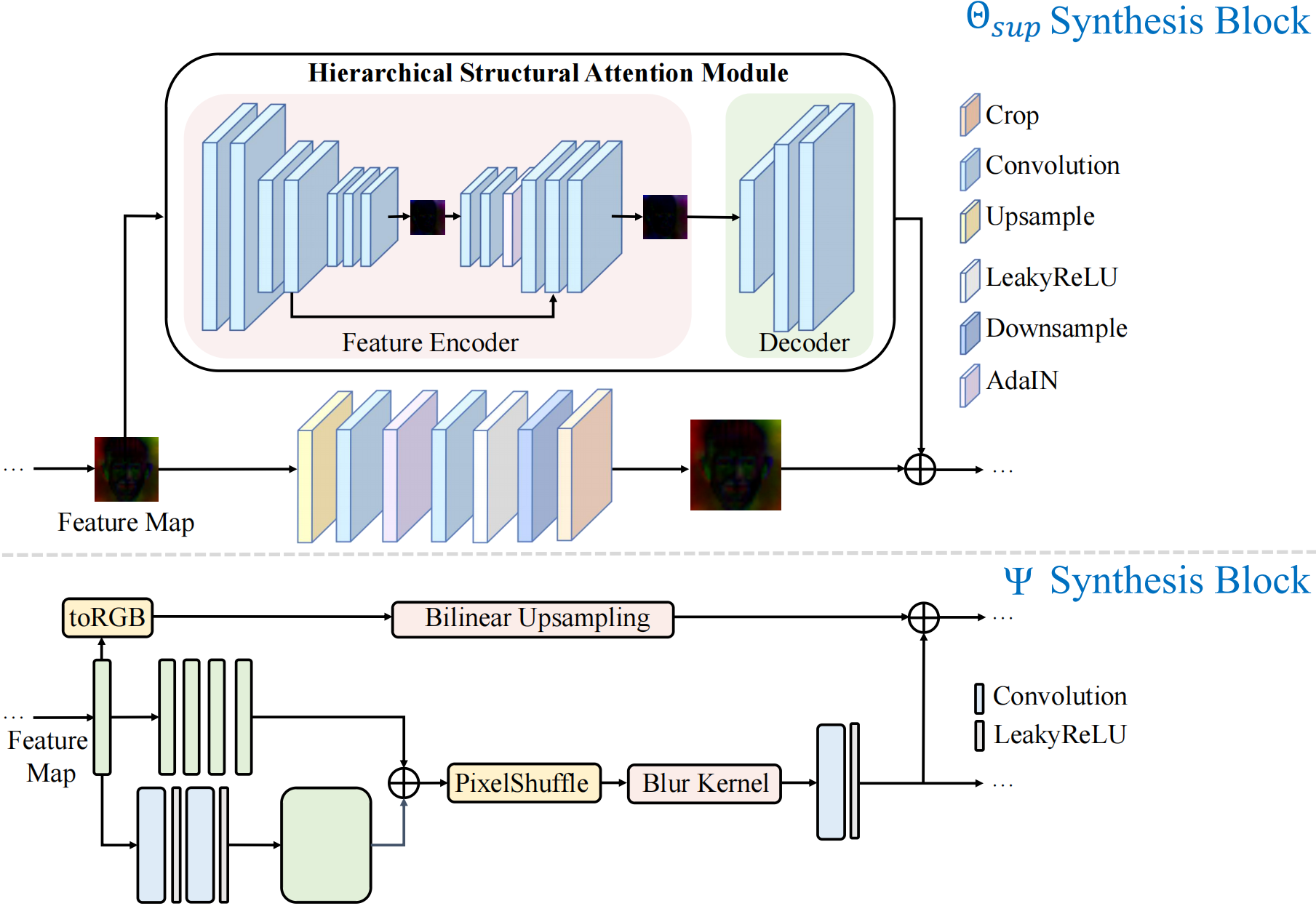}
   \caption{Synthesis block of super-resolution module $\Theta_{sup}$(Top) and neural rendering module $\Psi$(Bottom).}
   \label{fig:NeuralUpsampler}
\end{figure}
\section{Related Work}
\noindent \textbf{Learning-based Head Reconstruction, Generation and Control.}
The generation visual quality of Generative Adversarial Net(GAN)~\cite{goodfellow2020generative} is impressive. For 2D-based GANs, there are two main lines to control the head attributes. Since the well-designed StyleGAN~\cite{karras2019style} contains a semantically rich latent space, a commonly-used approach~\cite{shen2020interpreting,harkonen2020ganspace,shen2020interfacegan} is to explore and find “walking” directions that control a specific attribute of interest. Other works~\cite{deng2020disentangled,shoshan2021gan} propose to use contrastive learning~\cite{jaiswal2020survey}, collecting attribute-decoupled image pairs(different in only one attribute) to allow the model to learn the control of interested attributes. However, these methods rely on 2D-based generative models and thus the generated heads lack 3D view consistency.

NeRF~\cite{mildenhall2021nerf}-based generative models represent a 3D scene as a radiance field parameterized by a Multi-Layer Perceptron (MLP). pi-GAN~\cite{chan2021pi} has adopting a SIREN-based NeRF representation as the generator.  Because of the use of discriminant loss and the pure NeRF structure, pi-GAN cannot be trained at high resolution. GRAM~\cite{deng2022gram} propose to regulate point sampling and radiance field learning on 2D manifolds to enhance the image quality. EG3D~\cite{chan2022efficient} introduce a novel NeRF representation tri-plane and a dual-discriminator to improve the generation quality and rendering efficiency. These models improve the 3D consistency, but only randomly output heads and do not support head controllability. 

3D Morphable Model(3DMM)~\cite{blanz1999morphable} is a parametric face model which represent the face as a linear combination of a set of principle components derived from 3D scans via Principle Components Analysis (PCA). HeadNeRF~\cite{hong2022headnerf} is a NeRF-based reconstruction model that introduces 3DMM conditional priors. Its visual quality lacks realism and clarity, probably because it is essentially a decoder. At the same time, there are obvious discrepancy in expression, identity, hair contour and so on between the reconstructed head and the original head.

\noindent \textbf{Labeled Head Synthesis and Digital Head Engine.}
There are many human head features such as landmark coordinates, eye gaze angle, and hair color that are difficult to precisely label. Existing labeled database such as~\cite{le2012interactive,sagonas2016300,zhu2012face,jaiswal2013guided,sagonas2013semi,zhang2015appearance,kellnhofer2019gaze360} rely on experienced manual annotators, but due to the difficulty of collection, the data lacks geometry and appearance variation. This constraints the effectiveness of estimation models for head features. An increasing research~\cite{wood2021fake,wood2016learning,Bae_2023_WACV} seeks to replace or supplement real data with synthetic heads in some head-related tasks such as eye gaze estimation and face recognition, depending on graphics-based methods. The synthetic texture distribution however suffers a gap with real texture distribution and thus the method is far from satisfactory. Existing learning-based head models has a potential to address this issue, but they are not yet capable to generate a head consistent with the given label of the interested feature.

To address the above issues, and go one step further, we hope to propose a unified framework to fully realize the hyper-realistic head reconstruction, control, generation of heads consistent with the given label of the interested feature, and inversely estimate the head label. To the best of our knowledge, there is no similar work to our digital head engine MetaHead yet.

\section{Method}
In this section, we introduce how we design the hyper-realistic digital human engine MetaHead, including its reconstruction and synthesis framework controllable head radiance field(MetaHead-F), the loss terms, strategy of integrating the end-to-end decoder structure with pre-trained generator, and image generation framework LabelHead that generates heads consistent with the given customizable feature labels. The feature design space of MetaHead-F has many customizable options in addition to identity, illumination and texture. Since the expression prior is a necessary prerequisite in head decoupling control, here we take landmark feature as an example. Experiments illustrate that these four priors have enabled the model to achieve the state-of-the-art reconstruction and generation results. Examples of adding other features to the feature design space are given in \cref{sec:AttributeControllingResults} and \cref{fig:AttributeControllingResults} (d), (f)(please see the \href{https://ustc3dv.github.io/MetaHead/}{{project page}}  for video demos and more examples), which also verify that LabelHead allows MetaHead-F to precisely control more attributes such as gaze and hair color in addition to the common attributes.

\subsection{Controllable Head Radiance Field}
As shown in the left part of \cref{fig:pipeline}, MetaHead-F consists of HeadField and SemanticField. We set the conditional prior signals inputted to MetaHead-F to be optimizable, so that they could act as a bridge between HeadField and SemanticField, and the loss of SemanticField could exert semantic correction on HeadField through the optimization in latent space.

3DMM is just a PCA face model regressed on limited 3D scans, and thus the widely used 3DMM identity and expression priors do not capture the fine-grained shape or geometric details of real faces. For a given 2D head image, we pretrained an encoder to map the extracted feature with the face recognition model AdaFace~\cite{kim2022adaface} to a lower dimension 128, which acts as the 3D identity prior signal $\mathbf{z}_{id}$. 

For expression prior signal, we first randomly initialize the 3DMM coefficients and generate the corresponding face mesh. We specify the vertices on mesh corresponding to 68-point landmarks.  Then we project them onto the 2D head image, and optimize the 3DMM coefficients and the corresponding mesh by reducing the distance between the projection points and the ground truth landmark. 
After that, we project the optimized vertices to the 3D datum space by performing the inverse camera transformation, and denote the result vertices as 3D-HeadPoints, so as to guarantee that they does not contain camera pose information and can be decoupled from the latter. 
Furthermore, mesh (controlled by implicit 3DMM expression coefficient) cannot close eyes, but 3D-HeadPoints could have precise control on challenging expressions during training since it could be explicitly optimization.
Thus, we specify 3D-HeadPoints as expression prior signal $\mathbf{z}_{exp}$. 
In fact, we can incorporate more points as 3D-HeadPoints. 
Experiments illustrate that 68 points could already guarantee precise 3D reconstruction and control of expressions(see \cref{tab:Quantitative Ablation Study} and \cref{tab:QuantitiveComparison}). 

Besides, we input the 3DMM texture coefficient from the above optimization as the head texture condition signal $\mathbf{z}_{alb}$. Similarly, we regress to get 27-dimension spherical harmonic illumination coefficient as the head illumination condition signal $\mathbf{z}_{ill}$.

Given a 2D head image, first we randomly sample the points along the casted camera rays, denoted as $\mathbf{x}\in\mathbb{R}^{3}$, and perform positional encoding to it. The result $\gamma(\mathbf{x})$ is input to the MLP-based implicit neural function $h_{\theta}$:
\begin{equation}
        h_{\theta}: 
        (\gamma(\mathbf{x}), \mathbf{z}_{id}, \mathbf{z}_{exp},\mathbf{z}_{alb},\mathbf{z}_{ill}) \mapsto (\sigma, F),
        \label{equ:implicit neural function}
\end{equation}
where $\theta$ represents the network parameters, $\sigma$ is the density value at $\mathbf{x}$, and $F$ is an intermediate feature related to the radiance color at $\mathbf{x}$. 
After that, the final pixel color of feature map $I_{F} \in \mathbb{R}^{1024\times36\times36}$ is given by volume rendering:
 \begin{equation}
    \begin{split}
        I_F(r) & = \int_{0}^{\infty} w(t) \cdot F(r(t)) dt \\
        \textrm{where}~\quad w(t) & = exp(-\int_{0}^{t} \sigma(r(s))ds) \cdot \sigma(r(t)).
        \label{equ:volume render}
    \end{split}
\end{equation}
$t$ defines a sample point within near and far bounds and $r(t)$ represents a ray emitted from the camera center. The above structure are the volume rendering modules in  HeadField and SemanticField, denoting them as $\Phi_{i}$ for $i\in\left\{1,2\right\}$ respectively, and the corresponding feature maps are $ I_{F}^{(i)}$ for $i\in\left\{1,2\right\}$.

We designed a super-resolution module $\Theta_{sup}$ in HeadField. We then impose the conditional supervision signal into mapping network \textbf{M} and perform the result $\mathbf{w}$ to strengthen the 3D head prior signal in $\Theta_{sup}$. Through the combination of end-to-end decoder structure and GAN generator(pre-trained $\Theta_{sup}$), we greatly improved the visual quality of the output heads. 

Existing state-of-the-art face reconstruction and generation models suffer the dynamic-scene problem that the output head texture such as hair, eyebrow and teeth would visibly flickering during changing camera poses. To address this artifact, we design a hierarchical structural attention module(\cref{fig:NeuralUpsampler}) customized for 3D head vision tasks in each synthesis block of $\Theta_{sup}$. With the above design, we efficiently eliminate the structure distortion and 3D texture flickering(please watch the video demo in \href{https://ustc3dv.github.io/MetaHead/}{{project page}}). For SemanticField, the architecture of synthesis block of neural rendering upsampler $\Psi$ is shown in \cref{fig:NeuralUpsampler}. $\Theta_{sup}$ and $\Psi$ ensure that the resolution of the output heads $I_{1}$ and semantic mask $I_{2}$(a by-product) increases gradually, and the outputs could possess multi-view consistency.

\subsection{Field Learning \& Decoder-GAN Combination} 
The training loss functions include the $L_{1}$ loss $\mathcal{L}_{\textrm{photo}}$ and the perceptual LPIPS~\cite{zhang2018unreasonable} loss $ \mathcal{L}_{\textrm{perc}}$ between the reconstructed head $I_{1}$ and the real head $I_{\textrm{GT}}^{1}$, and the $L_{1}$ semantic loss $ \mathcal{L}_{\textrm{sem}}$ between the generated mask $I_{2}$ and the ground truth mask $I_{\textrm{GT}}^{2}$. We also impose an identity loss:
\begin{equation}
         \mathcal{L}_{\textrm{ID}} = 1 - \langle R(I_{1}),R(I_{\textrm{GT}}^{1})\rangle ,
        \label{equ:identity loss}
\end{equation}
where $R$ is an InsightFace~\cite{guo2021sample} face recognition network and $\langle\cdot,\cdot\rangle$ computes the cosine similarity between the arguments. Our proposed conditional supervision signals are inherently semantically decoupled and MetaHead-F would gradually learn the disentanglement control during optimization in latent space, which is guided by $\mathcal{L}_{\textrm{photo}}$ and $ \mathcal{L}_{\textrm{perc}}$, thus requiring no additional decoupling loss or contrastive learning. To preserve the visual attributes of the input heads, we minimize the $L_{2}$ norm of the optimized step in the feature space:
\begin{equation}
         \mathcal{L}_{\textrm{reg}} =  \sum\limits_{*} \lVert \mathbf{z}_{*}  - \mathbf{z}_{*}^0 \rVert_{2},
    \label{equ:regular loss}
\end{equation}
where $*$ stand for the four head prior conditions, and $\mathbf{z}_{*}^0$ denotes the original value.

\begin{figure*}[t]
  \centering
  \includegraphics[width=1\linewidth]{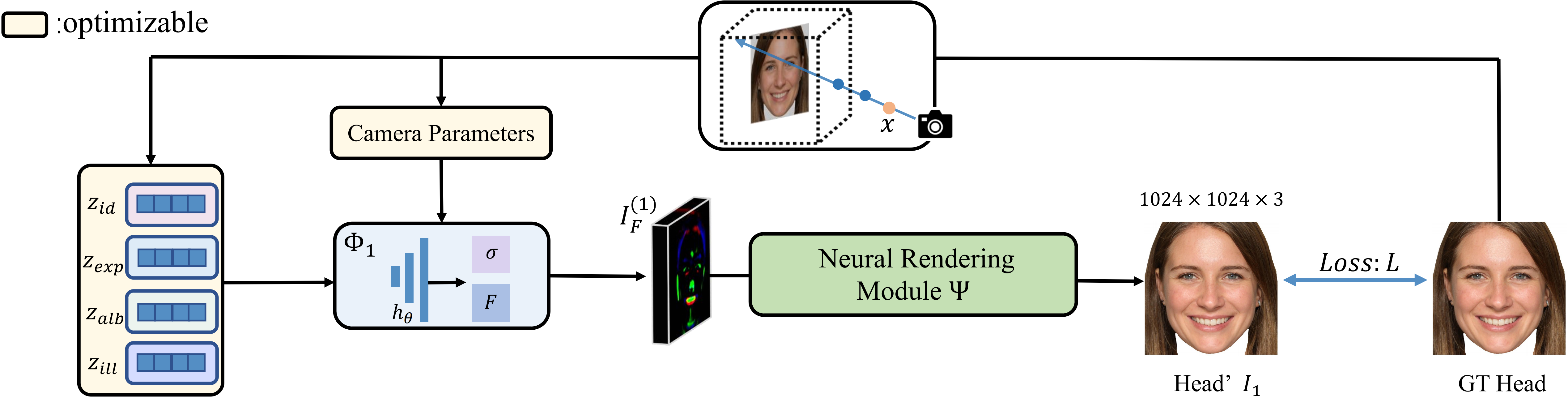}
   \caption{The model architecture of the baseline in ablation studies on MetaHead-F model designs. Our baseline removes the SemanticField and the super-resolution module $\Theta_{sup}$. In order to ensure that the baseline could output the same resolution as the HeadField, we replace $\Theta_{sup}$ with $\Psi$, which exactly raises the performance of the baseline. Meanwhile, we replace the identity and expression conditional supervisory signals $\mathbf{z}_{id}$ and $\mathbf{z}_{exp}$ with the traditional 3DMM fitting coefficients in baseline.}
   \label{fig:Baseline}
\end{figure*}
Previous work are usually a single GAN or decoder structure, because organizing the pre-trained generator after another network and training together would only lead to disordered signal transmission, thus generating mosaic-like images full of colored blocks. Our pre-trained generator $\Theta_{sup}$ requires an input latent feature $\mathbf{w}$ and a low-resolution Fourier feature. The interface of this Fourier feature makes it possible to combine the end-to-end structure and the generator. We analyzed the reason for the signal disorder, which is in that the distribution of input feature map $I_{F}^{(1)}$ is far from the pre-trained distribution of the generator. Based on this assumption, we design a multi-stage training strategy. In the first stage, we use $L_{2}$ loss denoted as $\mathcal{L}_{dist}$ to reduce the distribution deviation between the output distribution of $\Phi_{1}$ and the input distribution of pre-trained $\Theta_{sup}$. In the second stage, we reduce the weight of $\mathcal{L}_{dist}$ and the following formula is used to mildly replace the input feature map $M^{in}$:
\begin{equation}
        M^{in} = (1 -\lambda)\cdot\varphi + \lambda\cdot I_{F}^{(1)},
        \label{equ:replace feature}
\end{equation}
where $\varphi$ is a Fourier-distributed random feature and $\lambda$ is a weight parameter. 
\begin{figure*}[t]
  \centering
  \includegraphics[width=1\linewidth]{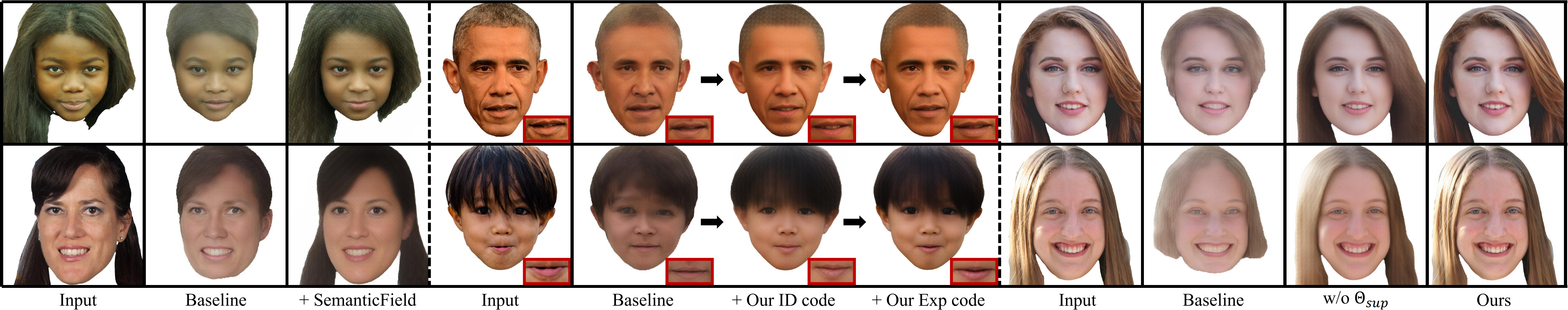}
   \caption{Qualitative ablation study on MetaHead-F model designs. Expressions are highlighted in red. Id: identity, Exp: expression. Refer to \cref{sec:AblationStudy} for details. }
   \label{fig:AblationStudy}
\end{figure*}

This simple yet effective method greatly improves the output visual quality of HeadField, and enables it to use real images as supervision, successfully combining the advantages of end-to-end training methods(supervised learning) and generative adversarial networks(unsupervised learning). Benefit from this, we bring forward an unparalleled GAN generation module HeadField with decoupled attributes control.

Therefore, the total learning objective is:
\begin{equation} 
\begin{split}
\mathcal{L}_{\textrm{total}} = &\lambda_{\textrm{photo}}\mathcal{L}_{\textrm{photo}} + \lambda_{\textrm{perc}}\mathcal{L}_{\textrm{perc}} + 
        \lambda_{\textrm{sem}}\mathcal{L}_{\textrm{sem}} + \\
        &\lambda_{\textrm{ID}}\mathcal{L}_{\textrm{ID}}+ \lambda_{\textrm{reg}}\mathcal{L}_{\textrm{reg}} +
        \lambda_{\textrm{dist}}\mathcal{L}_{\textrm{dist}},
        \label{equ:total loss}
        \end{split}
\end{equation}
where $\lambda_{*}$ are the loss weights.
\begin{table}
  \centering
  \resizebox{\linewidth}{!}{
  \begin{tabular}{|l|c|c|c|c|c|c|c|c|}
    \hline
    & \multicolumn{4}{c|}{CelebAMask-HQ~\cite{lee2020maskgan}} & \multicolumn{4}{c|}{FFHQ~\cite{karras2019style}} \\
    \hline
    Model & AED$\downarrow$~\cite{ren2021pirenderer} & ID$\uparrow$~\cite{chan2022efficient} & PSNR$\uparrow$ & SSIM$\uparrow$ & AED$\downarrow$ & ID$\uparrow$ & PSNR$\uparrow$ & SSIM$\uparrow$\\
    \hline\hline
    baseline  & 0.1103 & 0.5210 & 19.6 & 0.702  & 0.1009 & 0.5439 & 21.6 & 0.755 \\
    w/ SemanticField & $0.0939$ &  $0.5788$ & 22.85 & 0.8236 & $0.0826$ & $0.6051$ & 23.01 & 0.8607 \\
    w/ our identity code  & $0.1012$ & $\underline{0.8044}$ & $21.45$ & $\underline{0.8433}$  & $0.0908$ & $\underline{0.8667}$ & $22.08$ & $0.8878$ \\
    w/ our expression code  & $\underline{0.0355}$ & $0.5410$ & $21.80$ & $0.8218$ & $\underline{0.0222}$ & $0.5643$ & $22.48$ & $0.8751$ \\
    w/ $\Theta_{sup}$ & $0.0904$ & $0.5443$ & $\underline{28.43}$ & ${0.8429}$ & $0.0857$ & $0.5632$ & $\underline{29.70}$ & $\underline{0.9032}$\\
    \hline
    MetaHead-F & $\mathbf{0.0289}$ & $\mathbf{0.9058}$ & $\mathbf{29.98}$ & $\mathbf{0.8632}$ & $\mathbf{0.0150}$ & $\mathbf{0.9145}$ & $\mathbf{30.20}$ & $\mathbf{0.9247}$\\
    \hline
  \end{tabular}}
  \caption{Quantitative ablation study on model designs of MetaHead-F.}
  \label{tab:Quantitative Ablation Study}
\end{table}

\subsection{LabelHead: Top-down Head Image Synthesis}
Thanks to our controllable end-to-end head model MetaHead-F, and its huge feature design space, we can add any customizable head-related attributes to the feature design space, such as landmark(arbitrary number of points), eye gaze angle, pixel-wise semantic category~\cite{lee2020maskgan} and hair color. We train on image-label pairs. Then we freely edit the label value of interested attributes and MetaHead-F could generate a 3D controllable head which is consistent with the given label. Except for this, we can further fix one feature and manipulate other feature values, such as dimming the illumination and enlarging the pose, and thus to generate image-label paired training data under challenging scenes. LabelHead enables us to precisely 3D control more customizable attributes in addition to the common ones.
Besides, as shown in the right part of \cref{fig:pipeline}, we can impose the photometric loss($\mathcal{L}_{\textrm{photo}}$ and  $\mathcal{L}_{\textrm{perc}}$) to bottom-up fit the various attributes of interest of the input head, and estimate the label bidirectionally(see \cref{para:Attributes Fitting Using LabelHead}). This is unreachable for ~\cite{wood2021fake} under graphics-based forward rendering framework.

\begin{figure*}[t]
  \centering
   \includegraphics[width=1.0\linewidth]{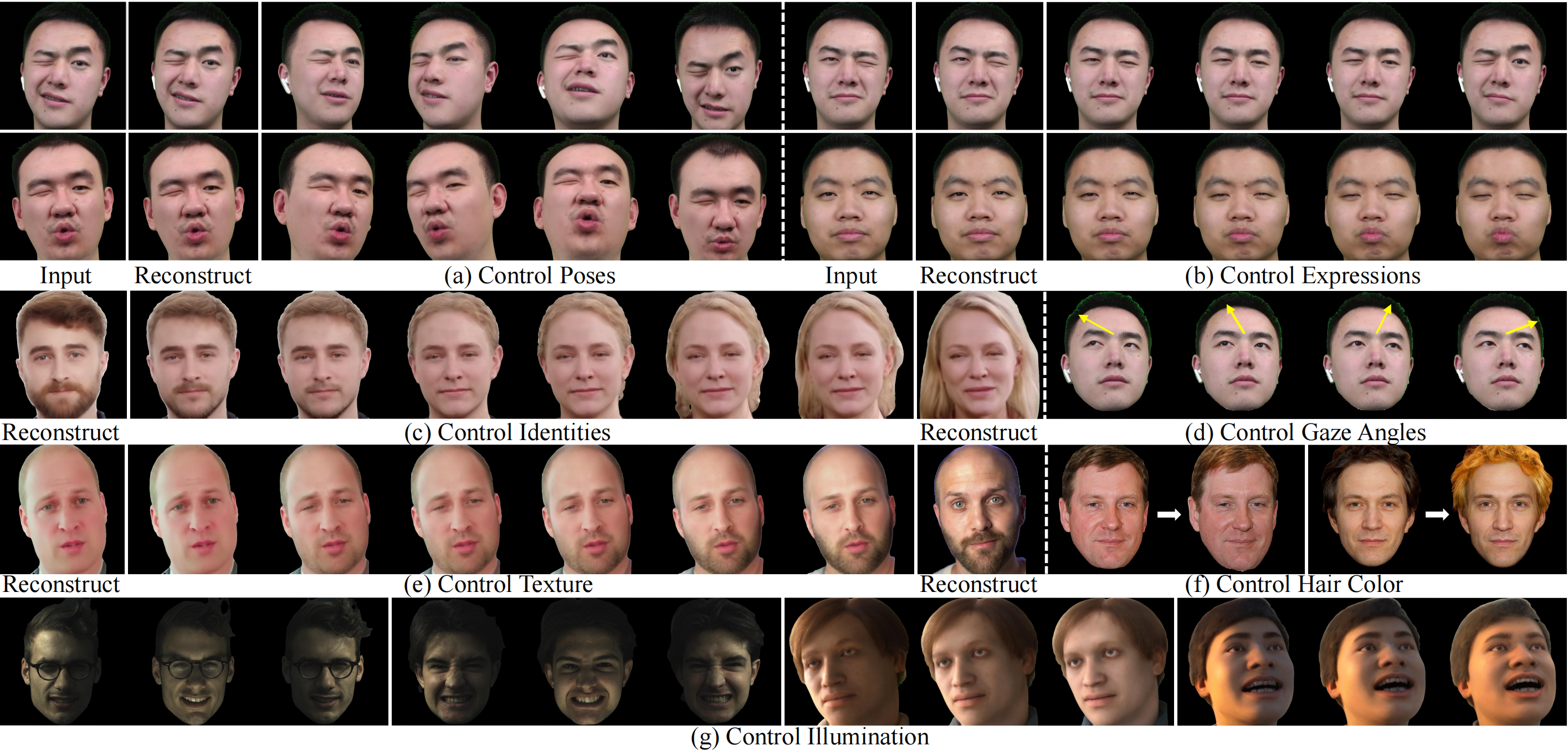}
   \caption{Attributes controlling results of MetaHead-F. (c) and (e) demonstrate progressive interpolation(from left to right) over the identity and texture of the reconstructions. MetaHead-F can achieve highly disentangled and precise 3D control over the identity, expression, texture, illumination, gaze angle, hair color, and camera pose of heads, and can also generate heads in a 3D-aware stylization manner.}
   \label{fig:AttributeControllingResults}
\end{figure*}
\begin{figure*}[t]
  \centering
   \includegraphics[width=1.0\linewidth]{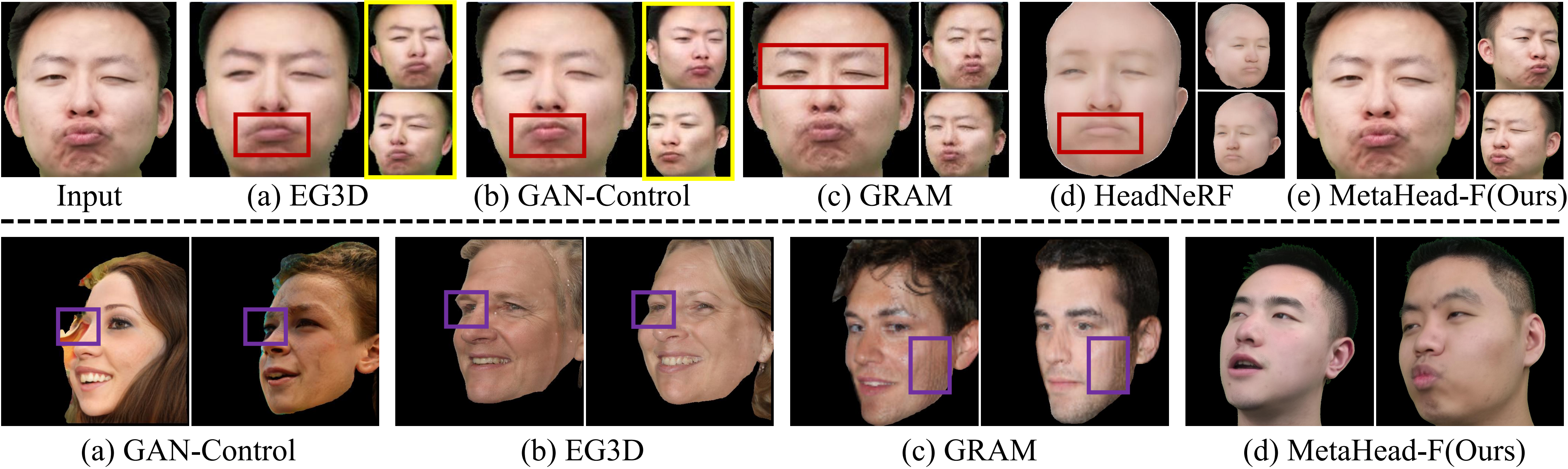}
   \caption{(Top)Reconstruction and attributes control accuracy comparison with existing methods. Red: expression inaccuracy. Yellow: identity and expression inconsistency. (Bottom)Heads generated by MetaHead-F in extreme poses, compared with existing methods. Previous methods generate heads with unreasonable artifacts such as blurry or missing eyes, and layered skin, while MetaHead-F produces robust results.
}
   \label{fig:MetaHead-F_Comparison}
\end{figure*}
\section{Experiments}
\label{sec:experiment}
\subsection{Implementation Details}
We train MetaHead-F on FFHQ dataset~\cite{karras2019style} and remain 6000 for reconstruction test. We also processed monocular videos of 500 people, and collected other videos containing exaggerated expressions and extreme poses with professional equipment as training data. For super-resolution module $\Theta_{sup}$, we pretrained it with the discriminator of StyleGAN2~\cite{karras2020analyzing} on FFHQ training data. The final model is trained for 120 hours with four 3090 GPUs.
\subsection{Ablation Study on MetaHead-F Model Designs}
\label{sec:AblationStudy}
We conduct ablation studies on FFHQ~\cite{karras2019style} testing data to validate the effectiveness of each component proposed in our head model MetaHead-F. \cref{fig:Baseline} illustrates the model architecture of the baseline, which shares the same training data and number of training epochs as MetaHead-F. Our baseline removes the SemanticField and the super-resolution module $\Theta_{sup}$. 
In order to ensure that the baseline could output the same resolution as the HeadField, we replace $\Theta_{sup}$ with $\Psi$, which exactly raises the performance of the baseline. 
Meanwhile, we replace the identity and expression conditional supervisory signals $\mathbf{z}_{id}$(identity code) and $\mathbf{z}_{exp}$(expression code) with the traditional 3DMM fitting coefficients in baseline. 
The reconstruction results are shown in \cref{fig:AblationStudy} and \cref{tab:Quantitative Ablation Study}. 
We add SemanticField to the baseline in the left part of \cref{fig:AblationStudy} and in row 4 of \cref{tab:Quantitative Ablation Study}. We replace 3DMM coefficients with our proposed conditional supervisory signals $\mathbf{z}_{id}$ and $\mathbf{z}_{exp}$ to the baseline in rows 5 and 6 of \cref{tab:Quantitative Ablation Study}, respectively. Besides, we replace 3DMM coefficients with our proposed signals one by one in the middle part of \cref{fig:AblationStudy}. 
Moreover, we replace $\Psi$ with $\Theta_{sup}$ in baseline in row 7 of \cref{tab:Quantitative Ablation Study}, and we replace $\Theta_{sup}$ in HeadField with $\Psi$ in the full model MetaHead-F in the right part of \cref{fig:AblationStudy}. 
ID~\cite{chan2022efficient} and AED~\cite{ren2021pirenderer} are common metrics to measure the identity and expression 3D reconstruction and control accuracy, respectively, calculated between the test and reconstructed heads. PSNR and SSIM are both visual quality metrics. 

The studies show that SemanticField precisely control the hair and mouth shape, our proposed identity and expression conditional prior signal enhance the control over fine-grained geometric details(identity and expression), and super-resolution module $\Theta_{sup}$ significantly improves the output visual quality. Ablation studies in dynamic scenes are given in the video demo(please see the \href{https://ustc3dv.github.io/MetaHead/}{{project page}}), which shows that $\Theta_{sup}$ and our proposed hierarchical structure attention module nicely solve the chronic problem of hair and teeth flickering when the viewing angle changes.
\begin{table}
  \centering
  \resizebox{\linewidth}{!}{
  \begin{tabular}{|c|c|c|c|c|c|c|}
    \hline
    Methods & Venue & FID$\downarrow$~\cite{heusel2017gans} & $DS_{\alpha}\uparrow$~\cite{deng2020disentangled} & $DS_{\beta}\uparrow$ & $DS_{\gamma}\uparrow$ & $DS_{\theta}\uparrow$\\
    \hline
    GIRAFFE~\cite{GIRAFFE}  & CVPR'21 & 32.6 & - & - & - & 36.2 \\
    pi-GAN~\cite{chan2021pi}  & CVPR'21 & 55.2 & - & - & - & 34.6 \\
    GRAM~\cite{deng2022gram} & CVPR'22 & 17.9 & - & - & - & 37.6 \\
    EG3D~\cite{chan2022efficient} & CVPR'22 & $\mathbf{4.7}$ & - & - & - & 39.2\\
    \hline
    DiscoFaceGAN~\cite{deng2020disentangled}  & CVPR'20 & 56.6 & 7.85 & 80.4 & 489 & 36.7 \\
    PIRenderer~\cite{ren2021pirenderer}  & ICCV'21 & 73.4 &  8.16 & 64.3 & - & 30.2 \\
    GAN-Control~\cite{shoshan2021gan}  & ICCV'21 & 14.6 & 9.26 & 69.2 & 496 & 39.8 \\
    HeadNeRF\cite{hong2022headnerf}  & CVPR'22 & 160.5 & 7.91 & 52.1 & 471 & 41.5 \\
    \hline
    MetaHead-F(Ours) & - & $\underline{6.7}$ & $\mathbf{15.52}$ & $\mathbf{93.3}$ & $\mathbf{511}$ & $\mathbf{47.2}$\\
    \hline
  \end{tabular}}
  \caption{Quantitative comparison of visual quality, and disentanglement and accuracy of attributes control with existing methods.}
  \label{tab:QuantitiveComparison}
\end{table}
\begin{figure*}[t]
  \centering
   \includegraphics[width=1.0\linewidth]{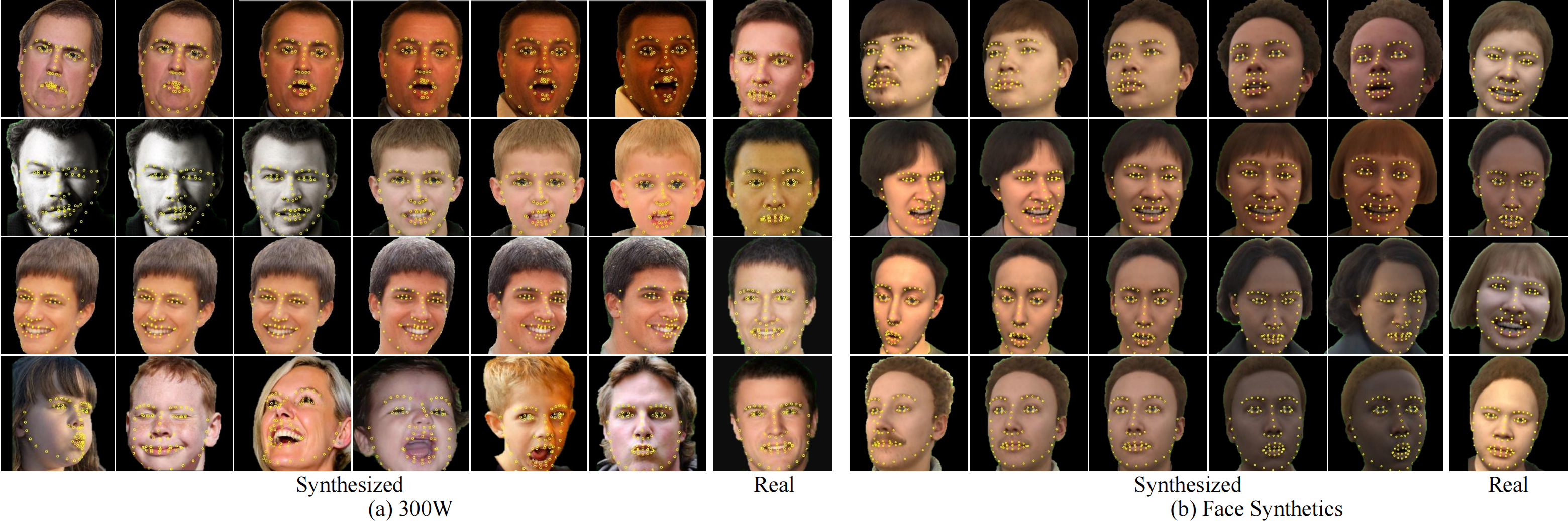}
   \caption{We fine-tuned MetaHead-F on \textbf{300W}~\cite{sagonas2016300} and Microsoft Face Synthetics~\cite{wood2021fake} respectively. Right side of sub-figures show real images. MetaHead-F synthesizes heads(left side of sub-figures) indistinguishable from real images with accurate feature(landmark) labels(marked in yellow) and much larger shape and appearance variation.}
   \label{fig:ShapeAndAppearanceVariation}
\end{figure*}
\begin{figure}[t]
  \centering
   \includegraphics[width=1.0\linewidth]{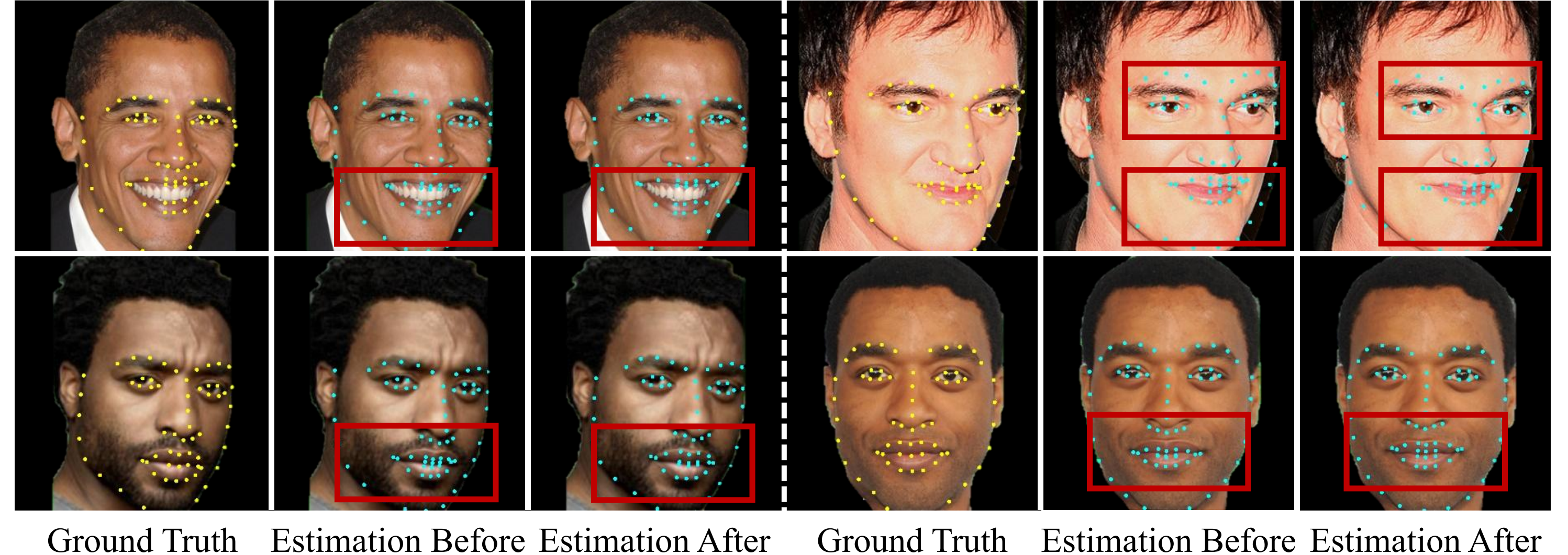}
   \caption{Landmark estimations on \textbf{300W}~\cite{sagonas2016300} testing data before and after adding the synthesized heads.}
   \label{fig:LandmarkEstimation}
\end{figure}
\vspace{2mm}
\subsection{Qualitative Evaluation on MetaHead-F}
\label{sec:Qualitative Evaluation on MetaHead-F}
\paragraph{Attributes Controlling Results}
\label{sec:AttributeControllingResults}
~\cref{fig:AttributeControllingResults} shows the reconstruction results of MetaHead-F and its separate control over identity, expression, texture, illumination, and camera poses. The last two showcases in (g) illustrate that MetaHead-F can stylizedly generate and control digital heads. As shown in \cref{fig:AttributeControllingResults}, MetaHead-F provides precise and view-consistent 3D control over the shape and appearance of heads in a highly decoupled manner, including challenging expressions and illumination. In comparison, previous models~\cite{chan2022efficient,deng2022gram,deng2020disentangled,shoshan2021gan} could only output or control moderate expressions. Furthermore, if we add the gaze angle, hair color and semantic label into the feature design space of MetaHead-F, MetaHead-F could further precisely control these three head features(see (d), (f) in  \cref{fig:AttributeControllingResults} and \cref{subsec:semantic label}), which cannot be achieved by any existing models.
\paragraph{Comparison on Reconstruction and Disentanglement} We present results on reconstruction accuracy and disentanglement of attributes control in row 1 of \cref{fig:MetaHead-F_Comparison}, comparing our method against four state-of-the-art head synthesis(~\cite{chan2022efficient,shoshan2021gan,deng2022gram}) or reconstruction-only(~\cite{hong2022headnerf}) methods. More examples are displayed in \cref{fig:RecontrunctionComparison} of the supplementary material. For each sub-figure, the left column corresponds to the reconstruction result, and the right column corresponds to the results of controlling camera poses. In \cref{fig:MetaHead-F_Comparison} row 1 (a)-(d), these four methods fail to reconstruct the right expression, and a few examples are highlighted in red. EG3D and GAN-Control fail to preserve the identity and expression consistency when only changing camera poses, which are highlighted in yellow. Besides, HeadNeRF missing the fine-grained appearance details such as chin wrinkle. In contrast, \cref{fig:MetaHead-F_Comparison} row 1 (e) shows that MetaHead-F precisely reconstructs all factors(covering shape and appearance) of the heads with highly consistent and precise pose control. Only poses change during the pose editing while other head properties remain unchanged, which shows a good and robust control disentanglement.
\paragraph{Comparison on Large Pose Generation} 
We present large pose generation results of GAN-Control~\cite{shoshan2021gan}, EG3D~\cite{chan2022efficient}, GRAM~\cite{deng2022gram} and ours in row 2 of \cref{fig:MetaHead-F_Comparison}. 
Previous methods fail to generate robust results, and suffer the unreasonable or blurry artifacts, which are highlighted in purple. 
A possible reason is that the data distribution of extreme poses in their training data is far from enough. 
Besides, GRAM produces layered artifacts in large pose, which is perhaps caused by the surface manifold learning.
In comparison, MetaHead-F could generate natural novel-view heads in the extreme camera pose.
\vspace{3mm}
\subsection{Quantitative Evaluation on MetaHead-F}
We compare our method to several state-of-the-art generative or reconstruction-only methods. GIRAFFE~\cite{GIRAFFE}, pi-GAN~\cite{chan2021pi}, GRAM~\cite{deng2022gram} and EG3D~\cite{chan2022efficient} are only able to control pose while DiscoFaceGAN~\cite{deng2020disentangled}, PIRenderer~\cite{ren2021pirenderer}, GAN-Control~\cite{shoshan2021gan} and HeadNeRF~\cite{hong2022headnerf} are controllable methods. 

We compare the visual quality of head generation using the Fr\`echet Inception Distance(FID) scores~\cite{heusel2017gans}(see column 3 of \cref{tab:QuantitiveComparison}). 
We use FFHQ~\cite{karras2019style} and CelebAMask-HQ~\cite{lee2020maskgan} as the real data, and measure FID scores between them and 50k randomly generated images.
The results show that our result is only second to EG3D. It is worth mentioning that since the output resolution of MetaHead-F is $1024$, to be fair to the above methods, we sacrificed to downsample the output to resolution of 512 with bilinear interpolation, which would result in the degradation of visual quality, and thus causing a higher FID score. Besides, EG3D couldn't control any attributes except for pose.
We compare the disentanglement and accuracy of attributes control on FFHQ~\cite{karras2019style} testing data using the Disentanglement Score(DS)(see columns 4-7 of \cref{tab:QuantitiveComparison}), which was proposed in ~\cite{deng2020disentangled}. $DS_{\alpha}$, $DS_{\beta}$, $DS_{\gamma}$ and $DS_{\theta}$ stand for DS score of identity, expression, illumination and pose, respectively. The results show that MetaHead-F achieves significantly better disentanglement and more precise control compared to existing methods.

\begin{figure}[t]
  \centering
   \includegraphics[width=1.0\linewidth]{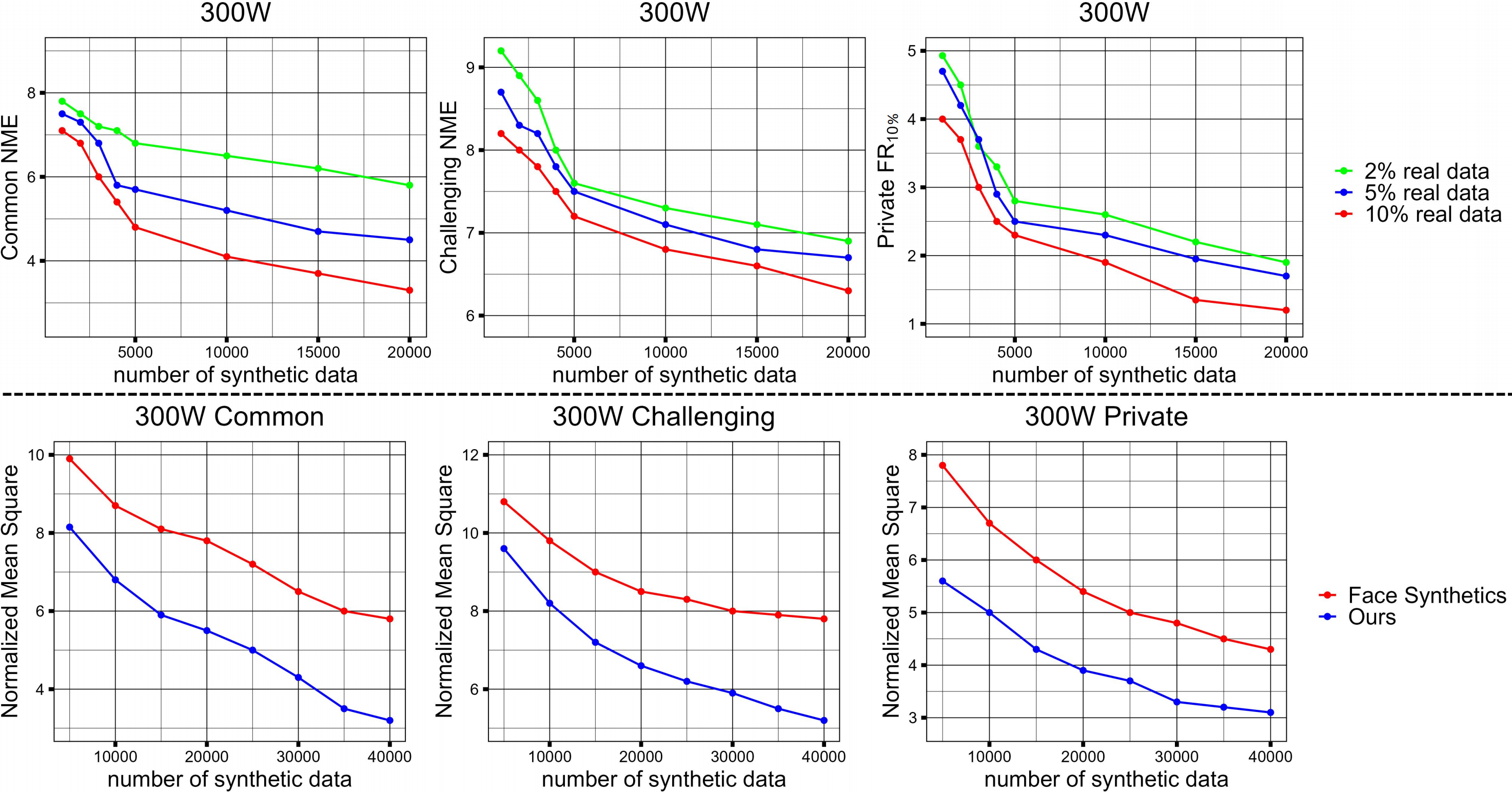}
   \caption{(Top) Our synthetic data significantly reduce the landmark estimation error. (Bottom) Landmark estimation error comparison between using state-of-the-art method(Face Synthetics Dataset~\cite{wood2021fake}) and our synthetic data.}
   \label{fig:ComparisonWithFaceSynthetics}
\end{figure}

\subsection{Qualitative Evaluation on LabelHead} 
\label{subsec:Qualitative Evaluation on LabelHead}
To equip the proposed MetaHead-F with the capability to generate heads consistent with the customizable head labels, we offer a generic top-down image generation framework LabelHead, which is able to be applied on any shape-appearance related head features. Due to limited space, we only discuss landmark feature in detail here, and experiment paradigm for other features are in the same way. We also discuss eye gaze angle and semantic label(category) feature quantitatively and qualitatively in \cref{subsec:eye gaze angle} and \cref{subsec:semantic label}, including synthesizing heads with consistent feature labels and the direct control of eye appearance and head shape.
\paragraph{Shape and Appearance Variation} 
We embed the landmark feature 3D-HeadPoints into the feature design space and fine-tune the pre-trained MetaHead-F on \textbf{300W}~\cite{sagonas2016300} training images for 2D face alignment task and Face Synthetics~\cite{wood2021fake} for 3D face alignment task, respectively, which have ground truth landmark labels. 
By providing a sequence of head attributes such as identity, illumination, texture, pose and 3D-HeadPoints as conditional signals, we can easily generate heads with diverse shape and appearance variation as shown in \cref{fig:ShapeAndAppearanceVariation}, while the latter is very important in label-annotated head generation. 
We then project 3D-HeadPoints onto the corresponding 2D images to obtain the landmark labels. 
We show 2D landmark labels in \cref{fig:ShapeAndAppearanceVariation} (a) rows 1-3, 3D landmark labels in \cref{fig:ShapeAndAppearanceVariation} (a) row 4 and \cref{fig:ShapeAndAppearanceVariation} (b).
\cref{fig:ShapeAndAppearanceVariation} (a) shows that LabelHead could generate super-realistic heads under challenging scenes with accurate labels. 
\cref{fig:ShapeAndAppearanceVariation} (b) shows that MetaHead-F could synthesize stylized heads with equally accuracy. 
Besides, \cref{fig:ShapeAndAppearanceVariation} also indicates that LabelHead can generate multi-task(2D and 3D) landmark labels with good generalization.
\paragraph{Qualitative Evaluation on Label-Estimation}
We first qualitatively evaluate whether synthesized heads help label estimation with small amount of real data. We train a landmark estimator using widely known backbone ResNet34~\cite{he2016deep} on image-landmark
pairs of \textbf{300W}~\cite{sagonas2016300} training data and test on \textbf{300W}~\cite{sagonas2016300} testing data. As shown in \cref{fig:LandmarkEstimation}, the model is under-fitted with purely real data which cause poor performance. However, after we add our generated heads which are 10 times larger than the amount of real data, the performance remarkably improves. Some accuracy comparison examples are highlighted in red in \cref{fig:LandmarkEstimation}. This
demonstrates that our synthesized heads indeed capture the correlation between landmark, head shape and head appearance,
and can be great useful for applications with small amount of real data.
\subsection{Quantitative Evaluation on LabelHead} 
\label{subsec:Quantitative Evaluation on LabelHead}
\paragraph{Quantitative Evaluation on Label-Estimation}
We also quantitatively evaluate whether our generated heads help label estimation with small amount of real data.  
The experiments are conducted on \textbf{300W}~\cite{sagonas2016300} following the common protocol in ~\cite{zhu2015face}, where we perform testing on three parts: the common, challenging and private subsets. 
The alignment accuracy is evaluated by the Normalized Mean Error(NME) and Failure Rate below a $10\%$ error threshold($\textrm{FR}_{10\%}$), while the lower means the better. 
 We train a landmark estimator using ResNet34~\cite{he2016deep} on image-landmark pairs of training data. Training data consists of k$\%$ of real images and n generated heads. \cref{fig:ComparisonWithFaceSynthetics} row 1 shows that as
we continuously add the synthetic heads, the performance keeps improving until saturation. 

Next we compare our generated data with the state-of-the-art landmark-labeled image generation method~\cite{wood2021fake} on head synthesis. 
Our synthesized heads and Microsoft Face Synthetics generated by ~\cite{wood2021fake} are used to train the landmark estimator ResNet34 respectively and the result models are tested on \textbf{300W} common, challenging and private subsets. 
Since the landmark labels are 3D in Microsoft Face Synthetics, we utilize InsightFace~\cite{guo2021sample} to perform a translation on jawline from 3D to 2D for fairness as guided in ~\cite{wood2021fake}. 
As shown in \cref{fig:ComparisonWithFaceSynthetics} row 2, our training data achieve superior results on each subset. The key reason is that MetaHead-F could generate super-realistic heads which is approximate to the distribution of real data and further cover the diverse challenging scenes.

\section{Application}
Due to the limited space, we demonstrate applications including one-shot facial retargeting, text-to-head generation and text-based 3D head manipulation in \cref{sec:Additional Application} in the supplementary material. 
\paragraph{Attributes Fitting Using LabelHead}
\label{para:Attributes Fitting Using LabelHead}
We take the fitting/estimation of the hair color feature as an example of the bottom-up label estimation here. 
We first embed the HSV(Hue, Saturation and Value) color values of hair as a feature into the feature design space of MetaHead-F and train MetaHead-F on FFHQ~\cite{karras2019style}. 
We then test on CelebAMask-HQ~\cite{lee2020maskgan}. We randomly initialize the label values of all features including hair color, then using the photometric loss between the test image and the generated head of MetaHead-F to optimize the label value, thus getting the HSV estimation.
\cref{tab:Attributes Fitting Using LabelHead} shows that the bottom-up fitting results are of high accuracy, which indicates that hair color feature captures the latent semantic position of hair and demonstrates
the effectiveness of the bidirectional label estimation using LabelHead.
\begin{table}[t]
  \centering
  \resizebox{\linewidth}{!}{
  \begin{tabular}{|l|c|c|c|c|}
    \hline
     & Hue(0-$360^{\circ}$) & Saturation(0-1) & Value(0-1) \\
    \hline
    Error(MSE~\cite{allen1971mean})  & 0.113 & 0.049 & 0.011\\
    \hline
  \end{tabular}}
  \caption{Bottom-up estimation error of the hair HSV color values using LabelHead.}
  \label{tab:Attributes Fitting Using LabelHead}
\end{table}

\section{Conclusion}
We have presented a digital head engine MetaHead, which consists of a controllable head radiance field(MetaHead-F) to super-realistically generate or reconstruct view-consistent controllable digital heads, enabling much more precise and decoupled 3D controllability over 3D identity, expression, texture, illumination and pose of the generated heads than existing state-of-the-art methods, and a generic top-down image generation framework LabelHead to generate heads consistent with the given customizable feature labels, which also enables MetaHead-F to control any shape-appearance related head features and bidirectionally estimate the labels of head features.

\bibliographystyle{ieee_fullname}
\bibliography{egbib}

\clearpage
\begin{strip}
\centering
\Large{\textbf{Supplementary Material}}
\end{strip}
\appendix
In this supplementary material, we provide both qualitative and quantitative results that were not included in our main manuscript. In order to demonstrate our dynamic results, we also provide the video demos, which can be found in the project page: \href{https://ustc3dv.github.io/MetaHead/}{https://ustc3dv.github.io/MetaHead/}.

\begin{figure*}[htbp]
  \centering
  \includegraphics[width=1\linewidth]{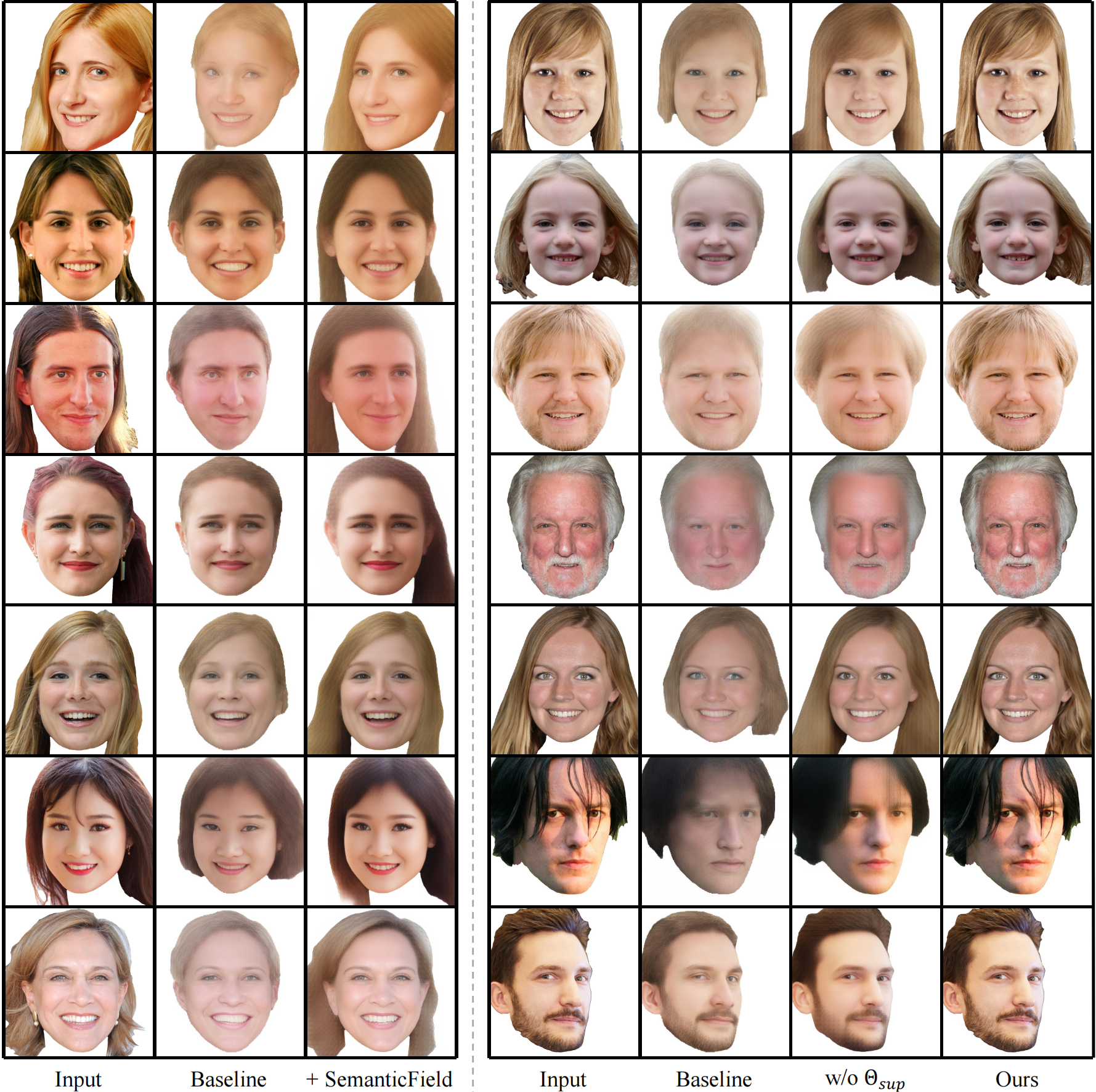}
   \caption{Qualitative ablation study on MetaHead-F model designs for SemanticField and the super-resolution module $\Theta_{sup}$. Please refer to \cref{sec:AblationStudy} of the main manuscript for details.}
   \label{fig:AblationStudy_1}
\end{figure*}
\begin{figure*}[htbp]
  \centering
  \includegraphics[width=1\linewidth]{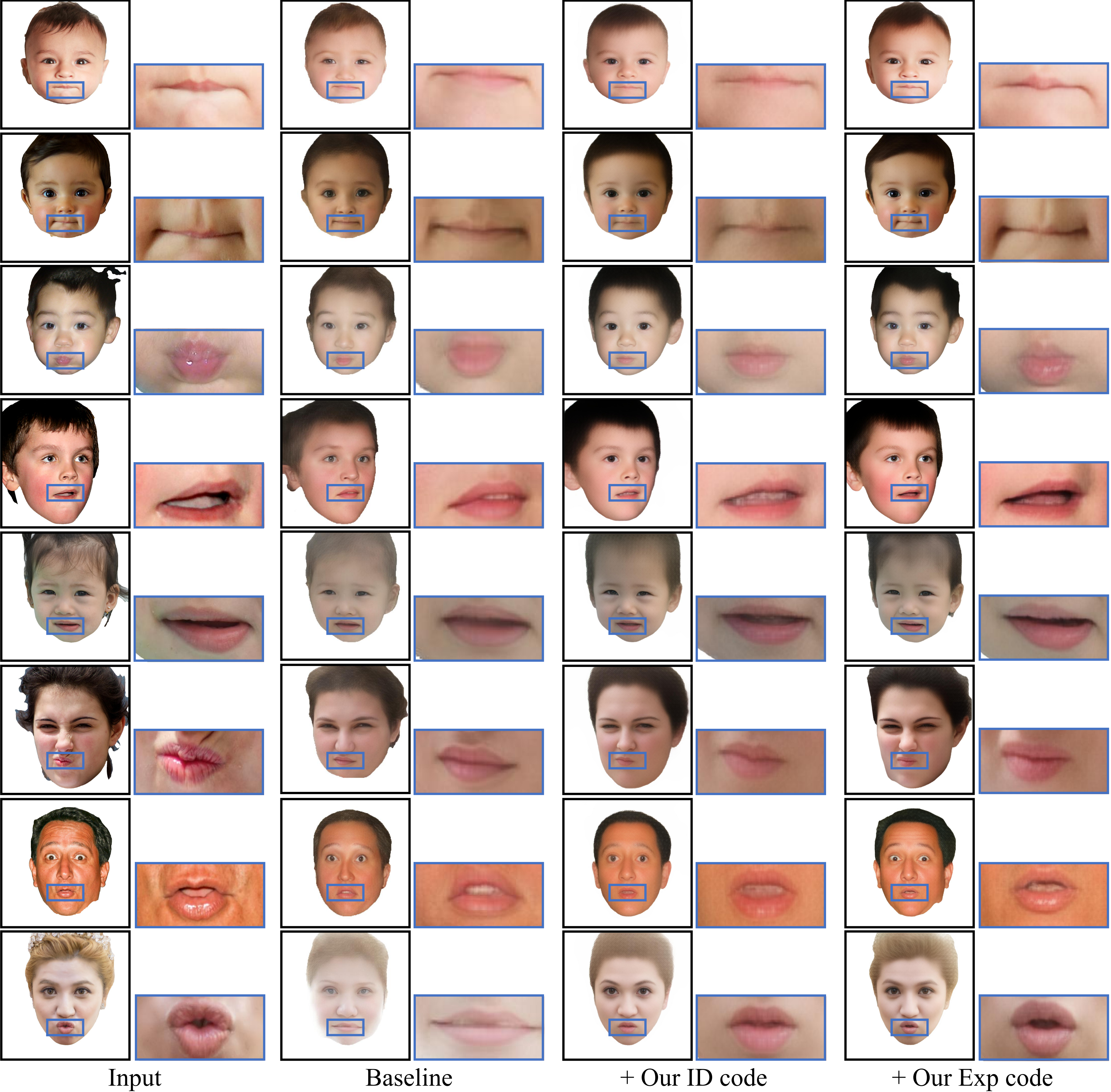}
   \caption{Qualitative ablation study on MetaHead-F model designs for our proposed conditional supervision head signals. Please refer to \cref{sec:AblationStudy} of the main manuscript for details. Expressions are highlighted in blue. Id: identity, Exp: expression.}
   \label{fig:AblationStudy_2}
\end{figure*}
\begin{figure*}[htbp]
  \centering
  \includegraphics[width=1\linewidth]{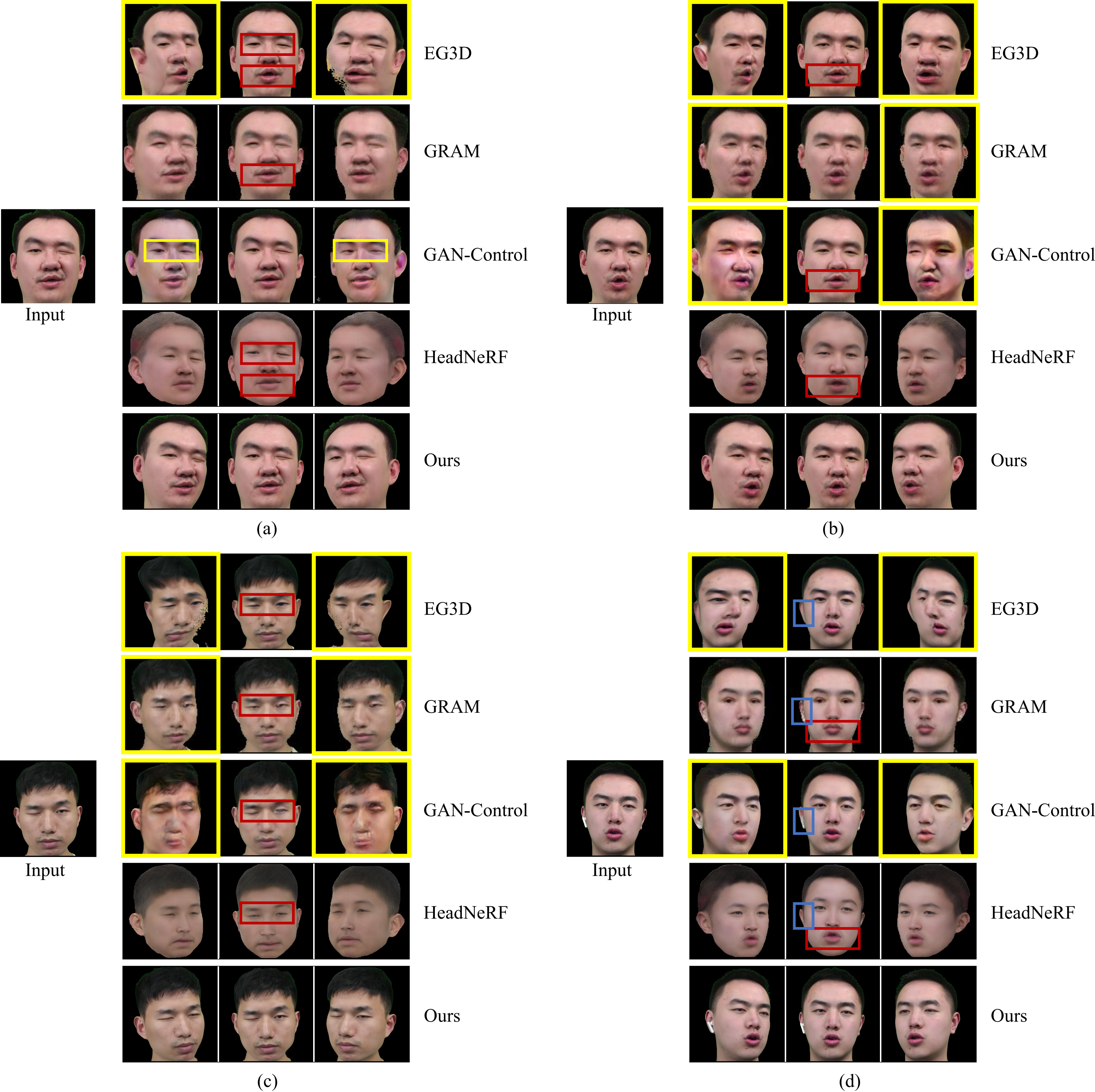}
   \caption{Reconstruction and attributes control accuracy comparison with existing methods. For each sub-figure, column 3 corresponds to the reconstruction results, while columns 2 and 4 correspond to the results obtained by controlling the poses of the reconstructions. Red: expression inaccuracy. Yellow: identity and expression inconsistency. Blue: head accessories inaccuracy. Please refer to \cref{para:Additional Comparison on Reconstruction} for details.}
   \label{fig:RecontrunctionComparison}
   \vspace{-2mm}
\end{figure*}
\begin{figure*}[htbp]
  \centering
  \includegraphics[width=1\linewidth]{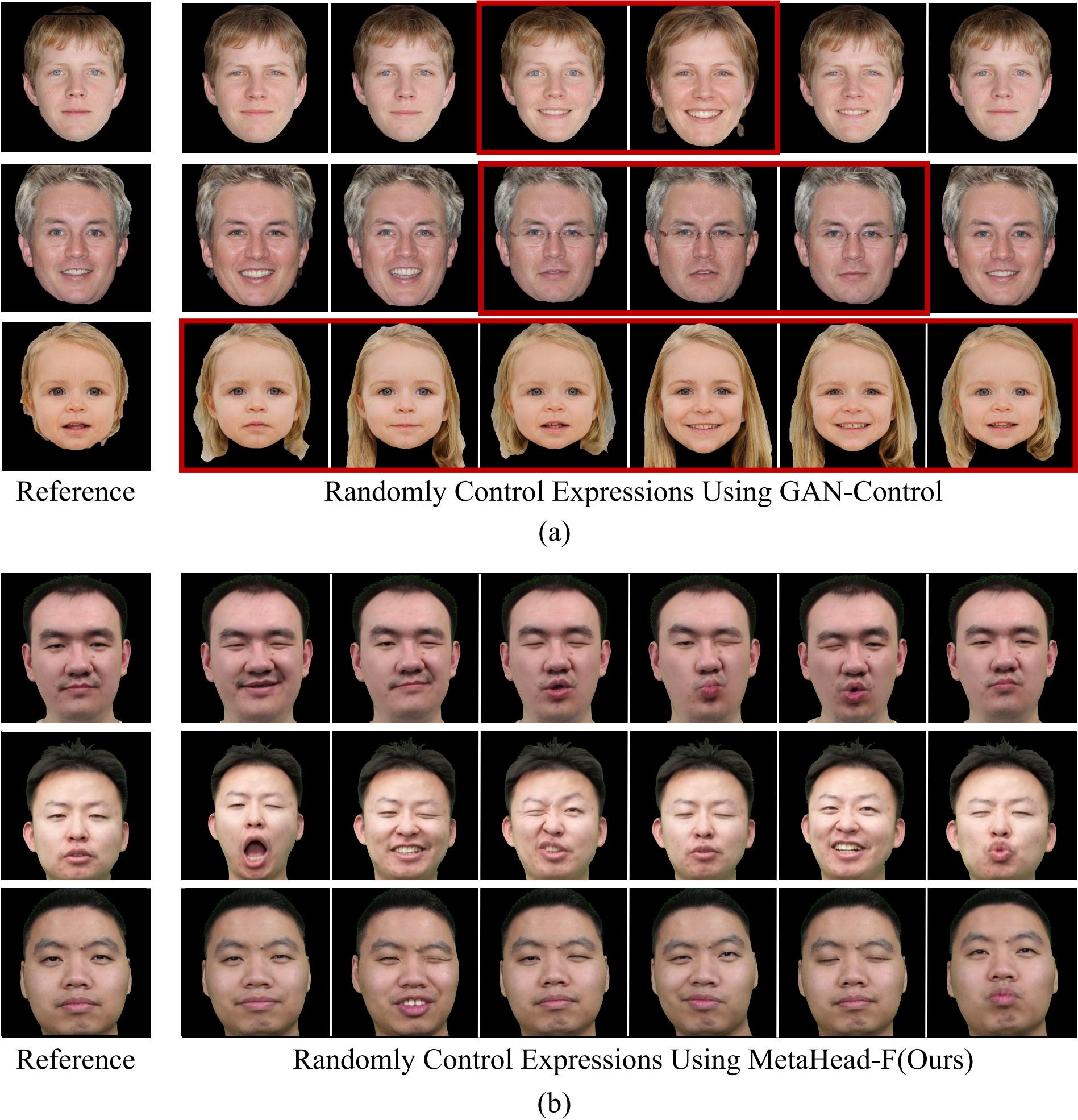}
   \caption{(a)Heads generated by GAN-Control~\cite{shoshan2021gan} with different expressions. Each row is generated
with the same parameters except the expression code. Some examples where GAN-Control produces identity inconsistencies when only the expression is supposed to change are highlighted in red. (b)For a reference head, MetaHead-F randomly control its expression in a highly consistent and photo-realistic manner.}
   \label{fig:Disentanglement}
   \vspace{-2mm}
\end{figure*}
\begin{figure*}[htbp]
  \centering
  \includegraphics[width=1\linewidth]{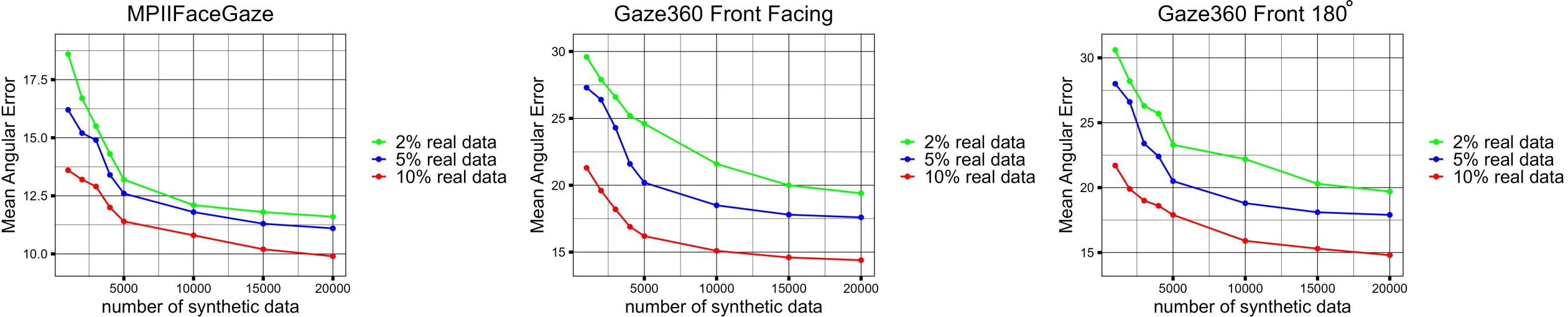}
   \caption{Performance comparison on \textbf{Gaze360}~\cite{gaze360_2019} and  \textbf{MPIIFaceGaze}~\cite{zhang2017s} using real and synthetic data. Our synthetic data significantly reduce the gaze estimation error.}
   \label{fig:Gaze}
\end{figure*}
\begin{figure*}[htbp]
  \centering
  \includegraphics[width=1\linewidth]{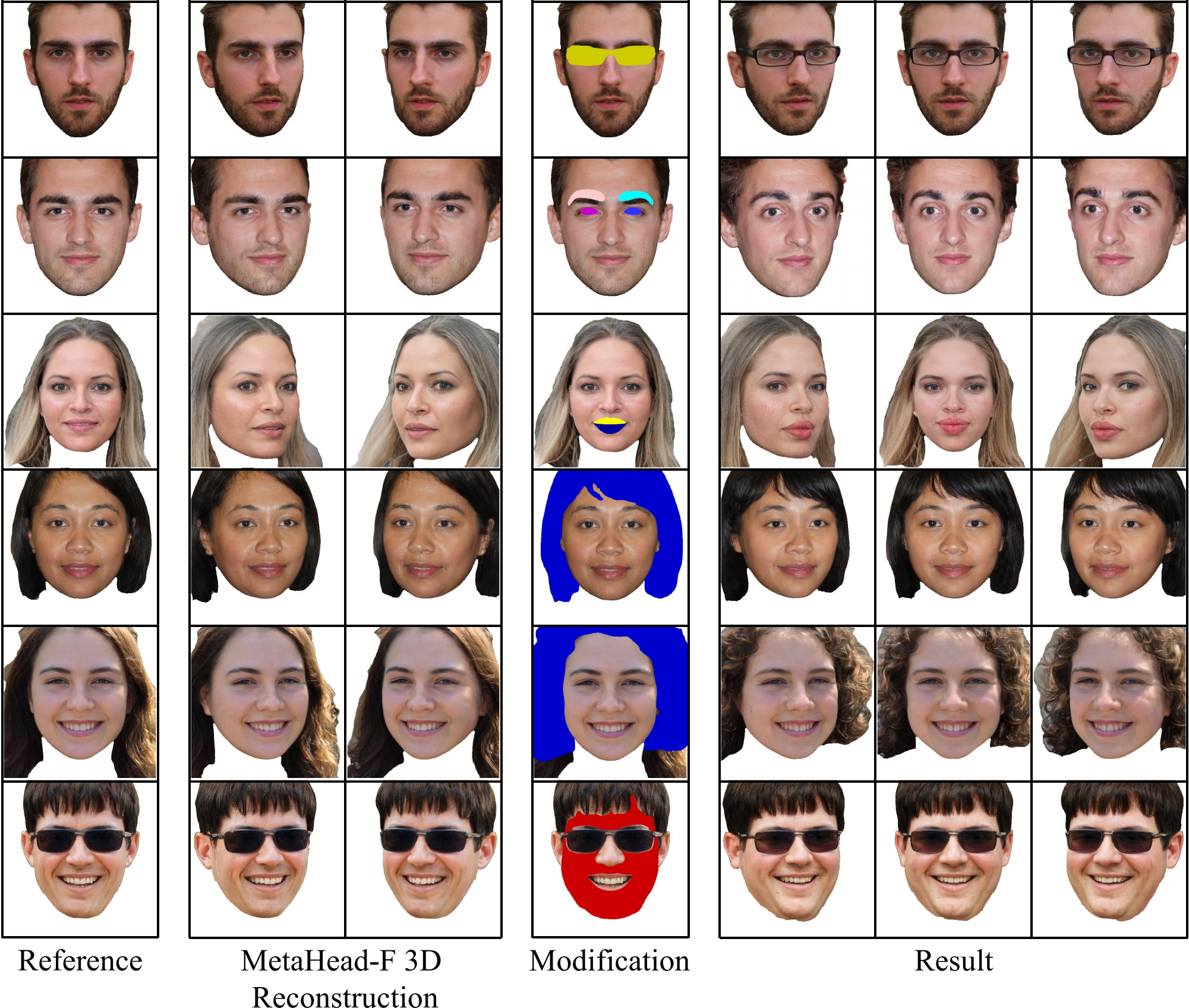}
   \caption{Results of local head editing using semantic labels. LabelHead allows users to perform locally 3D fine-grained head shape control and editing in a disentangled and view-consistent manner.
   Please refer to ~\cref{para:Head Shape Control Using Semantic Labels} for details.}
   \label{fig:Semantic}
\end{figure*}
\begin{figure*}[htbp]
  \centering
  \includegraphics[width=1.0\linewidth]{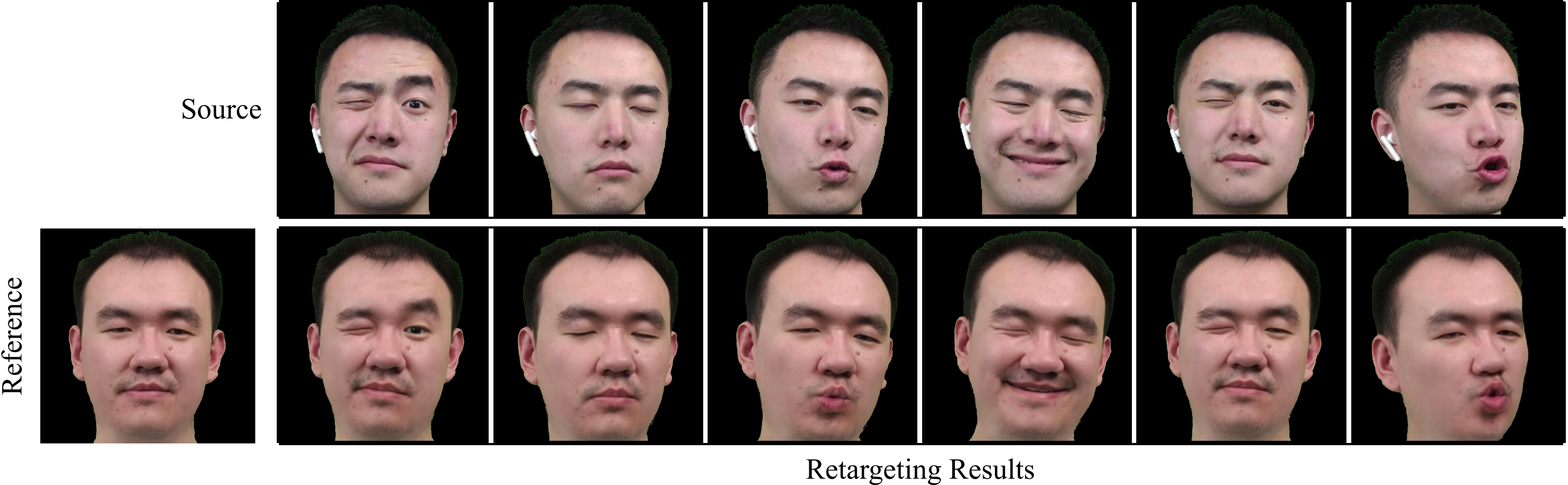}
   \caption{Application of head model MetaHead-F: one-shot facial retargeting.}
   \label{fig:Driven}
\end{figure*}
\begin{figure*}[t]
  \centering
  \includegraphics[width=1\linewidth]{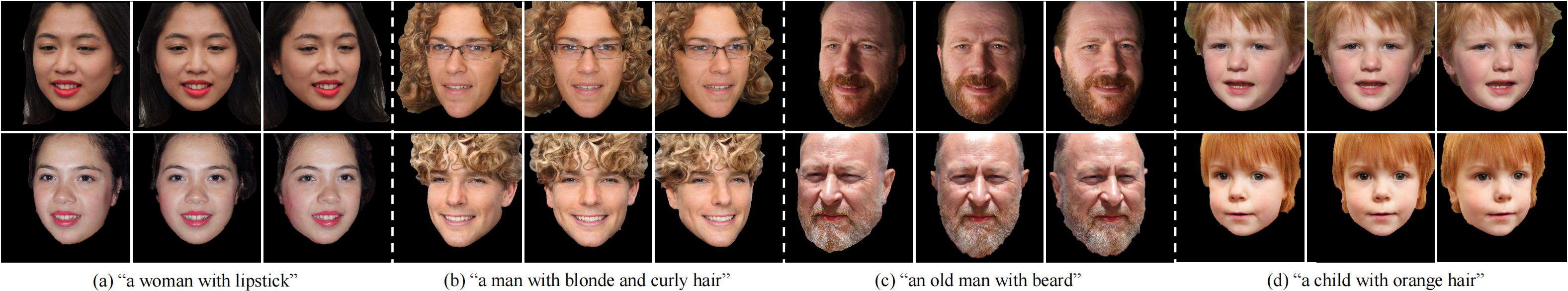}
   \caption{Application of head model MetaHead-F: text-to-head generation.}
   \label{fig:TextGenerate}
\end{figure*}
\begin{figure*}[!ht]
  \centering
  \includegraphics[width=1\linewidth]{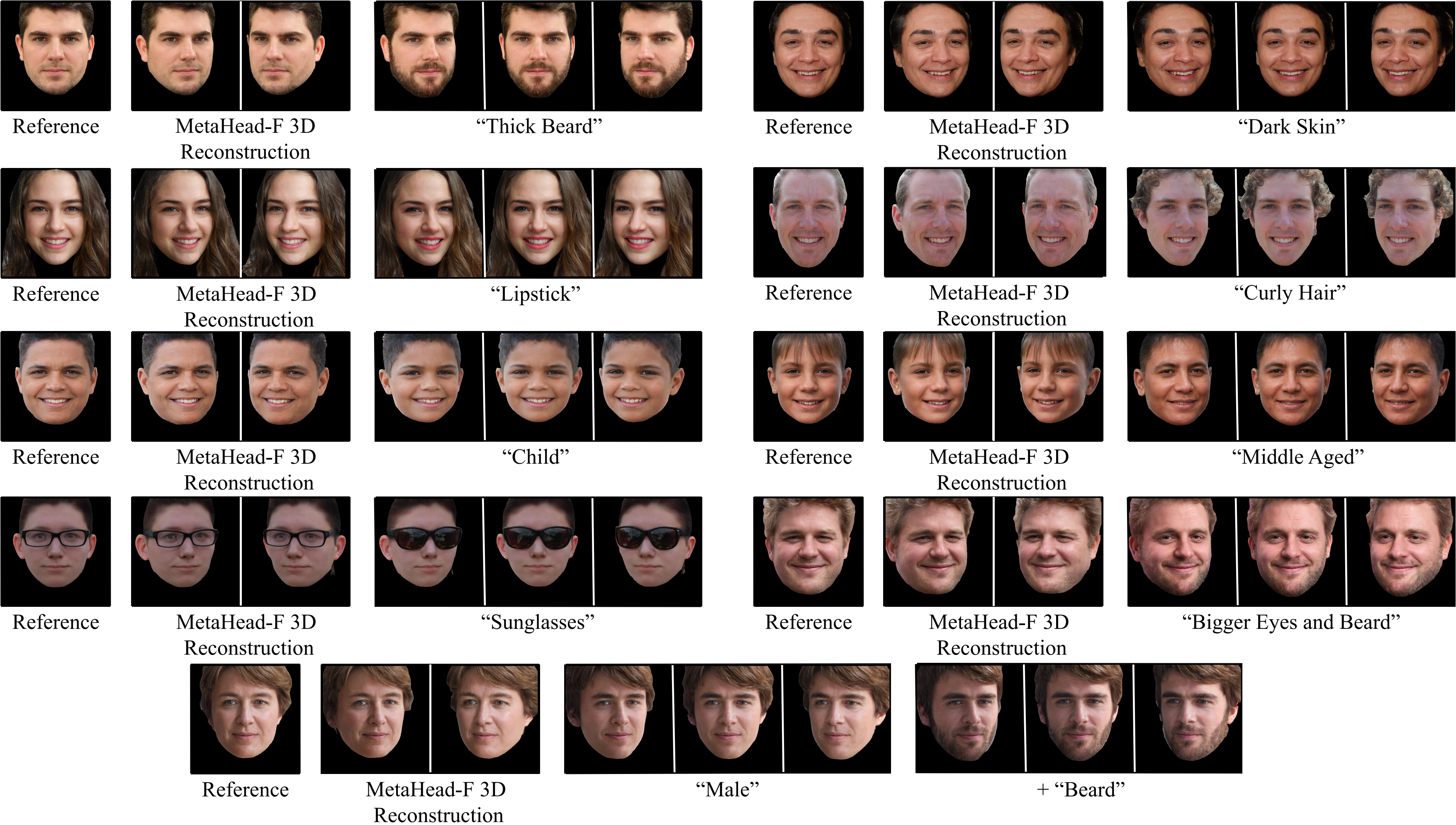}
   \caption{Application of head model MetaHead-F: text-based 3D head manipulation.}
   \label{fig:TextEdit}
\end{figure*}
\section{Additional Experiments}
\subsection{Ablation Study on MetaHead-F model designs}
\label{sec:Additional Qualitative Results}
\paragraph{Additional Qualitative Ablation Studies}
\label{sec:Additional Qualitative Ablation Studies}
We show more ablation experiments on SemanticField, super-resolution module $\Theta_{sup}$ and our proposed conditional supervision head signals in \cref{fig:AblationStudy_1} and \cref{fig:AblationStudy_2}. The experimental settings are the same as in \cref{sec:AblationStudy} of the main manuscript. Studies show that SemanticField precisely control the hair and mouth shape, our proposed identity and expression conditional prior signal enhance the control over fine-grained geometric details(identity and expression), and super-resolution module $\Theta_{sup}$ significantly improves the output visual quality.

\textbf{We strongly
recommend to watch the video demos in the \href{https://ustc3dv.github.io/MetaHead/}{{project page}}}, which contains ablation studies on the super-resolution module $\Theta_{sup}$ with our proposed hierarchical structural attention module to solve head texture flickering problem in dynamic scenes and the qualitative comparison with the existing state-of-the-art head models.
\subsection{Qualitative Evaluation on MetaHead-F}
\paragraph{Additional Comparison on Reconstruction and Control Accuracy}
\label{para:Additional Comparison on Reconstruction}
We present more results on reconstruction and attributes control accuracy in \cref{fig:RecontrunctionComparison}, comparing our method against four state-of-the-art head synthesis(~\cite{chan2022efficient,shoshan2021gan,deng2022gram}) or reconstruction-only(~\cite{hong2022headnerf}) methods. For each sub-figure, column 3 corresponds to the reconstruction results, while columns 2 and 4 correspond to the results obtained by controlling the camera poses of the reconstructions.
In \cref{fig:RecontrunctionComparison}, these four methods fail to reconstruct the right expression, and a few examples are highlighted in red. 
EG3D~\cite{chan2022efficient}, GRAM~\cite{deng2022gram} and GAN-Control~\cite{shoshan2021gan} fail to preserve the identity and expression consistency when only changing camera poses, which are highlighted in yellow. 
Besides, HeadNeRF~\cite{hong2022headnerf} missing the fine-grained appearance details such as acne and moles. 
The above four models fail to render the details of head accessories such as headphones, which are highlighted in blue.

In contrast, \cref{fig:RecontrunctionComparison} shows that MetaHead-F precisely reconstructs all factors(covering shape and appearance) of the heads with highly view-consistent and precise pose control. During pose editing, only the poses change while the other head properties remain unchanged. This demonstrates effective and robust control disentanglement.

\paragraph{Additional Comparison on Expression Control Disentanglement}
We present more results on the expression control, comparing our method against the state-of-the-art controllable head synthesis method GAN-Control~\cite{shoshan2021gan}. ~\cref{fig:Disentanglement} shows the generated heads using GAN-Control and our method MetaHead-F respectively. For each sub-figure, column 1 shows a reference
image, columns 2 to 7 show images generated with random expression control. Each row corresponds to the same person.

In ~\cref{fig:Disentanglement} (a), we can see that GAN-Control fails to preserve the identity when changing only the facial expression code. A few examples are
highlighted in red. Moreover, we observe that with the original range of the expression parameters,
the model results in only a small variation of expressions. 

~\cref{fig:Disentanglement} (b) shows that MetaHead-F generates compelling photo-realistic heads with highly
consistent, precise expression control. 
When controlling expressions for the heads in each row, the other head attributes, such as identity and texture, remain unchanged.
Besides, MetaHead-F could enable diverse expression variation including frowning, pouting, curling lips, etc.
\begin{table}
  \centering
  \resizebox{50mm}{!}{
  \begin{tabular}{|c|c|c|}
    \hline
    Methods & Venue & ID$\uparrow$~\cite{chan2022efficient}\\
    \hline
    GIRAFFE~\cite{GIRAFFE}  & CVPR'21 & 0.64 \\
    pi-GAN~\cite{chan2021pi}  & CVPR'21 & 0.67 \\
    GRAM~\cite{deng2022gram} & CVPR'22 & 0.74 \\
    EG3D~\cite{chan2022efficient} & CVPR'22 & 0.77 \\
    \hline
    DiscoFaceGAN~\cite{deng2020disentangled}  & CVPR'20 & 0.62 \\
    PIRenderer~\cite{ren2021pirenderer}  & ICCV'21 & 0.64\\
    GAN-Control~\cite{shoshan2021gan}  & ICCV'21 & 0.66\\
    HeadNeRF\cite{hong2022headnerf}  & CVPR'22 & 0.71\\
    \hline
    Ours & - & $\mathbf{0.79}$\\
    \hline
  \end{tabular}}
  \caption{Quantitative comparison between MetaHead-F and existing methods in terms of 3D view consistency.}
  \label{tab:view-consistency}
\end{table}
\subsection{Additional Quantitative Evaluation on MetaHead-F}
We quantitatively evaluate the 3D view consistency assessed by multi-view facial identity consistency(ID), following the method proposed in ~\cite{chan2022efficient}. We calculate the mean Arcface~\cite{deng2018arcface} cosine similarity score between pairs of views of the same synthesized head rendered from two random camera poses. ~\cref{tab:view-consistency} reports the comparison results on the FFHQ~\cite{karras2019style} testing data. Our method achieves the state-of-the-art view consistency.
\subsection{Additional Quantitative Evaluation on LabelHead}
\label{subsec:eye gaze angle}
Due to limited space, we only discuss landmark feature in detail in the main manuscript. We discuss synthesizing heads with consistent eye gaze labels here and exhibit the direct control of eye appearance using gaze label in the video demo(please see the \href{https://ustc3dv.github.io/MetaHead/}{{project page}}).
\paragraph{Quantitative Evaluation on Label-Estimation}
In order to better understand humans – their desires, intents and states of mind – one need to be able to observe and perceive certain behavioral cues. Eye gaze direction is one such cue: it is a strong form of non-verbal communication, signalling engagement, interest and attention during social interactions. 
In recent years, methods for gaze estimation have not yet reached a satisfactory level of performance. This is primarily due to the lack of sufficiently large and diverse labeled training data for the task.
Collecting precise and highly varied gaze data with ground truth, particularly outside of the lab, is a challenging task.

We embed the eye gaze angle as the gaze feature into the feature design space of MetaHead-F, and fine-tune the pre-trained MetaHead-F on the training images of two widely used gaze datasets \textbf{Gaze360}~\cite{gaze360_2019} and \textbf{MPIIFaceGaze}~\cite{zhang2017s} respectively, which include ground truth gaze labels.

By providing a sequence of head attributes such as identity, illumination, texture, pose and eye gaze angle as conditional signals, we can easily generate heads, which are consistent with the gaze angle label, with diverse shape and appearance variation, covering the diverse challenging scenes.

We quantitatively evaluate whether our generated heads help label estimation with small amount of real data. First, the experiment is conducted on \textbf{Gaze360}~\cite{gaze360_2019}.
We train an appearance-based gaze estimator Pinball LSTM~\cite{gaze360_2019} on image-gaze pairs of training data and test on \textbf{Gaze360}~\cite{gaze360_2019} testing data. Training data consists of k$\%$ of real images and n generated heads.

The estimation accuracy is evaluated by the mean angular error which are provided separately for samples where the subject is looking within $90^{\circ}$(Front $180^{\circ}$) and $20^{\circ}$(Front facing) of the camera direction.  \cref{fig:Gaze} shows that as we continuously add the synthetic heads, the performance of the estimator keeps improving until saturation. Experiments on \textbf{MPIIFaceGaze}~\cite{zhang2017s} also provide similar results(see \cref{fig:Gaze}). This suggests that our synthetic heads indeed capture the correlation between gaze, eye shape, and eye appearance, and are helpful for applications with little real data or less well annotated data.

Besides, since we embed the gaze angle feature into the feature design space, MetaHead-F could precisely control eye gaze angle(direction). We demonstrate dynamic control results on eye gaze in the video demo(please see the \href{https://ustc3dv.github.io/MetaHead/}{{project page}}).
\subsection{Additional Qualitative Evaluation on LabelHead}
\label{subsec:semantic label}
Apart from the landmark and gaze features, we also discuss synthesizing heads with consistent semantic labels and the control of head shape using semantic labels here.
\paragraph{Head Shape Control Using Semantic Labels}
\label{para:Head Shape Control Using Semantic Labels}
As is classified in CelebAMask-HQ~\cite{lee2020maskgan}, each pixel in the head semantic mask falls into 19 distinct categories including skin, eyebrows, ears, mouth, lip, etc. Given a head image, we flatten the semantic category corresponding to each pixel into a latent representation as the semantic label. We then embed the semantic label feature into the feature design space of MetaHead-F. 

We fine-tune the pre-trained head model MetaHead-F on CelebAMask-HQ~\cite{lee2020maskgan}, in which each image has a segmentation mask of 19-class facial attributes. After training, given the reference head image, we randomly initialize the label values of all features including semantic label, then using the photometric loss between the test image and the generated head of MetaHead-F to optimize the label value, thus getting the semantic mask estimation. 

We conduct interactive head shape manipulation by locally editing(drawing) on the obtained semantic mask and leverage MetaHead-F to generate the corresponding modified free-view 3D heads. As is shown in \cref{fig:Semantic}, the head shape control using semantic labels is disentangled and view-consistent.

\section{Additional Applications}
\label{sec:Additional Application}
\paragraph{One-shot Facial Retargeting}
\cref{fig:Driven} shows the application
of MetaHead-F for one-shot facial retargeting. Our expression and identity condition head signals are decoupled and can precisely control the head. Therefore, for a source subject's video or monocular images, we extract and replace the identity signals of the source subject with that of the test reference head. Then, the pre-trained MetaHead-F could perform high-fidelity single-view
3D reconstruction and facial retargeting(reenactment). 

\paragraph{Text-to-head Generation}
A natural way to customize 3D heads is to use language guidance. However, discovering semantically meaningful latent manipulations usually requires painstaking human examination of the many degrees of freedom. We leverage the power of Contrastive Language-Image Pre-training (CLIP)~\cite{radford2021learning} models in order to develop a text-based interface for MetaHead-F head generation and manipulation that does not require such manual effort. Our simple yet efficient approach for leveraging CLIP to guide image generation is to perform the direct latent code optimization with the CLIP loss:
\begin{equation}
    \mathcal{L}_\text{CLIP}(\mathbf{z}) = \Dclip(M(\mathbf{z}), t),
\end{equation}
where $M$ is the pre-trained MetaHead-F, $t$ is the text prompt, and $\Dclip$ is the cosine distance between its image and text CLIP embedding arguments. As shown in \cref{fig:TextGenerate}, one can finely customize the super-realistic 3D heads using text descriptions. 

\paragraph{Text-based 3D Head Manipulation}
We can further semantically 3D edit generated heads using text prompts through solving the optimization problem of minimize the cosine distance between the CLIP embeddings of its text and image inputs. Similarity to the input head is enforced by the identity loss:
\begin{equation}
         \mathcal{L}_{\textrm{ID}}(\mathbf{z}) = 1 - \langle R(M(\mathbf{z})),R(M(\mathbf{z}_{s}))\rangle,
        \label{equ:CLIP identity loss}
\end{equation}
where $\mathbf{z}_{s}$ is the source latent code, $R$ is an InsightFace~\cite{guo2021sample} face recognition network and $\langle\cdot,\cdot\rangle$ computes the cosine similarity between the arguments. 
As shown in \cref{fig:TextEdit}, we can achieve a wide variety of disentangled, meaningful and view-consistent 3D head control faithful to the text prompt.

\section{Discussion}
\paragraph{Ethics Statement}
Our digital head engine MetaHead focuses on technical development. Our approach can be used to super-realistically generate or reconstruct view-consistent 3D controllable digital heads, generate digital heads consistent with the given customizable feature labels and bidirectionally estimate the labels of features such as landmark coordinates, eye gaze angle, hair color and so on that are difficult to annotate before. 
However, since our digital head engine could generate heads at a quality that some might find difficult to differentiate from real human heads, we believe that it is essential to develop safeguarding measures to mitigate the potential for misuse. 

\end{document}